\documentclass{article}

\usepackage{microtype}
\usepackage{graphicx}
\usepackage{subfigure}
\usepackage{booktabs} 

\usepackage{hyperref}

\usepackage[accepted]{icml2025}

\usepackage{amsmath}
\usepackage{amssymb}
\usepackage{mathtools}
\usepackage{amsthm}

\usepackage{natbib}
\usepackage{wrapfig}
\usepackage{float}
\usepackage{multirow}
\usepackage{bbding}

\usepackage[capitalize,noabbrev]{cleveref}

\theoremstyle{plain}

\theoremstyle{definition}

\theoremstyle{remark}

\usepackage[textsize=tiny]{todonotes}

\icmltitlerunning{Learning Adaptive Lighting via Channel-Aware Guidance}

\begin{document}

\twocolumn[
\icmltitle{Learning Adaptive Lighting via Channel-Aware Guidance}



\icmlsetsymbol{equal}{*}

\begin{icmlauthorlist}
\icmlauthor{Qirui Yang}{equal,tju,vivo}
\icmlauthor{Peng-Tao Jiang}{equal,vivo}
\icmlauthor{Hao Zhang}{vivo}
\icmlauthor{Jinwei Chen}{vivo}
\icmlauthor{Bo Li}{vivo}
\icmlauthor{Huanjing Yue}{tju}
\icmlauthor{Jingyu Yang}{tju}
\end{icmlauthorlist}

\icmlaffiliation{tju}{Tianjin University, Tianjin, China.}
\icmlaffiliation{vivo}{vivo Mobile Communication Co., Ltd, Hangzhou, China}

\icmlcorrespondingauthor{Peng-Tao Jiang}{pt.jiang@vivo.com}
\icmlcorrespondingauthor{Jingyu Yang}{yjy@tju.edu.cn}

\icmlkeywords{Machine Learning, ICML}

\vskip 0.3in
]



\printAffiliationsAndNotice{\icmlEqualContribution} 

\begin{abstract}
\label{abs}
Learning lighting adaptation is a crucial step in achieving good visual perception and supporting downstream vision tasks. Current research often addresses individual light-related challenges, such as high dynamic range imaging and exposure correction, in isolation. However, we identify shared fundamental properties across these tasks:
i) different color channels have different light properties, and ii) the channel differences reflected in the spatial and frequency domains are different. Leveraging these insights, we introduce the channel-aware Learning Adaptive Lighting Network (LALNet), a multi-task framework designed to handle multiple light-related tasks efficiently. Specifically, LALNet incorporates color-separated features that highlight the unique light properties of each color channel, integrated with traditional color-mixed features by Light Guided Attention (LGA). The LGA utilizes color-separated features to guide color-mixed features focusing on channel differences and ensuring visual consistency across all channels. Additionally, LALNet employs dual domain channel modulation for generating color-separated features and a mixed channel modulation and light state space module for producing color-mixed features. Extensive experiments on four representative light-related tasks demonstrate that LALNet significantly outperforms state-of-the-art methods on benchmark tests and requires fewer computational resources. 
\textit{We provide an online demo at \href{https://xxxxxx2025.github.io/LALNet/}{LALNet}.}
\end{abstract}

\vspace{-0.6cm}
\section{Introduction}
\vspace{-0.1cm}
\label{intro}

\begin{figure}[ht]
    \centering
    \includegraphics[width=0.99\linewidth]{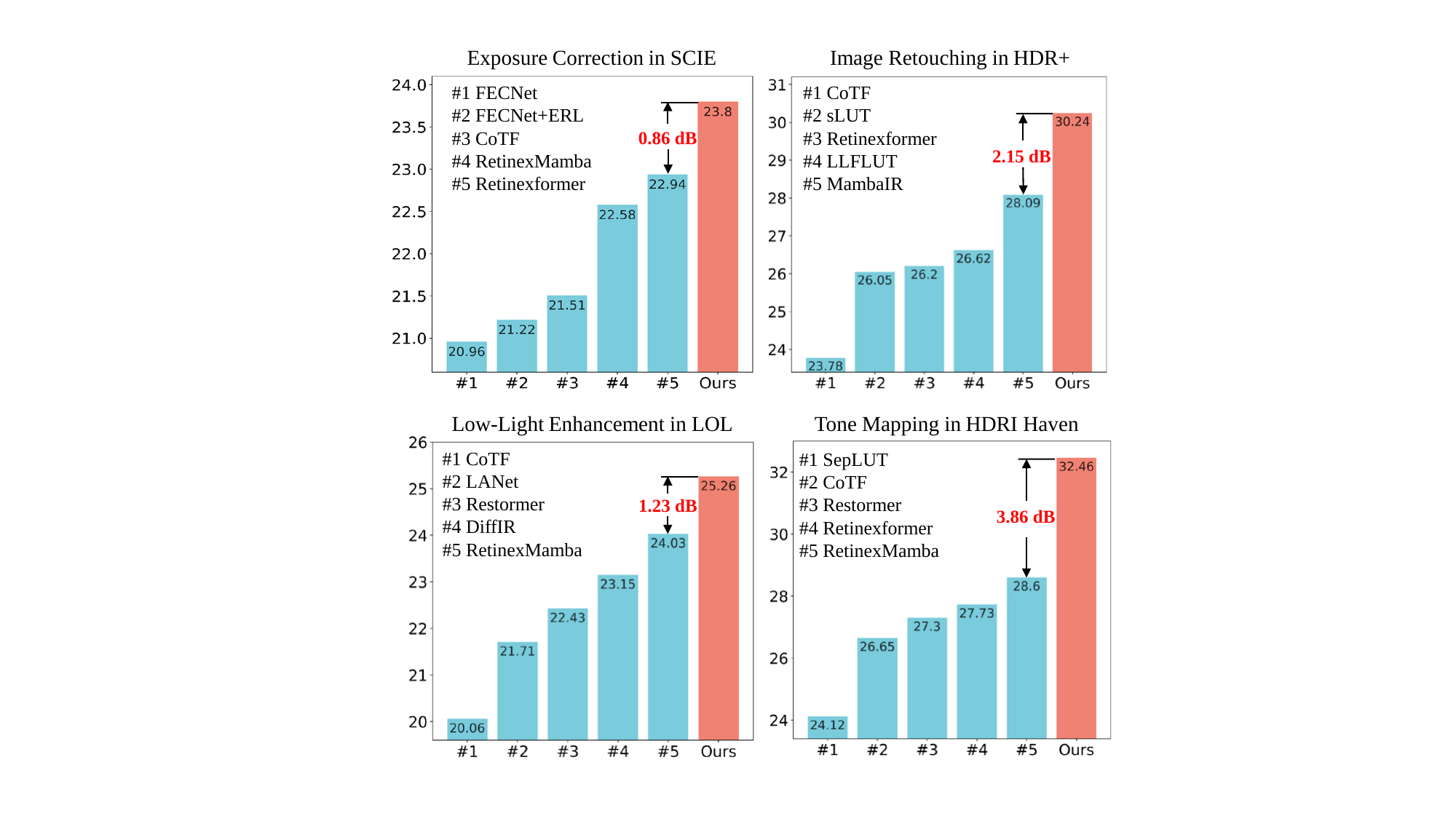}
    \vspace{-0.3cm}
    \caption {Our LALNet significantly outperforms state-of-the-art methods on four representative benchmark tests of light-related image enhancement, including image retouching, tone mapping, low-light enhancement, and exposure correction.}
    \vspace{-0.4cm}
    \label{examples}
\end{figure}

Photography is the art of light. 
The quality of an image is crucial for effective visual presentation and robust performance in subsequent computer vision tasks.
However, images taken under poor lighting conditions often exhibit degraded quality, which not only affects visual presentation but also poses challenges for tasks such as object detection and tracking. 
Consequently, learning adaptive lighting has emerged as a pivotal step in achieving robust visual perception and supporting downstream vision tasks.
This process is analogous to the perception of the human visual system, that is, light adaptation, which enables us to maintain stable visual perception across diverse lighting environments.

\begin{figure*}[t]
    \centering	  
    \centering{\includegraphics[width=0.98\textwidth]{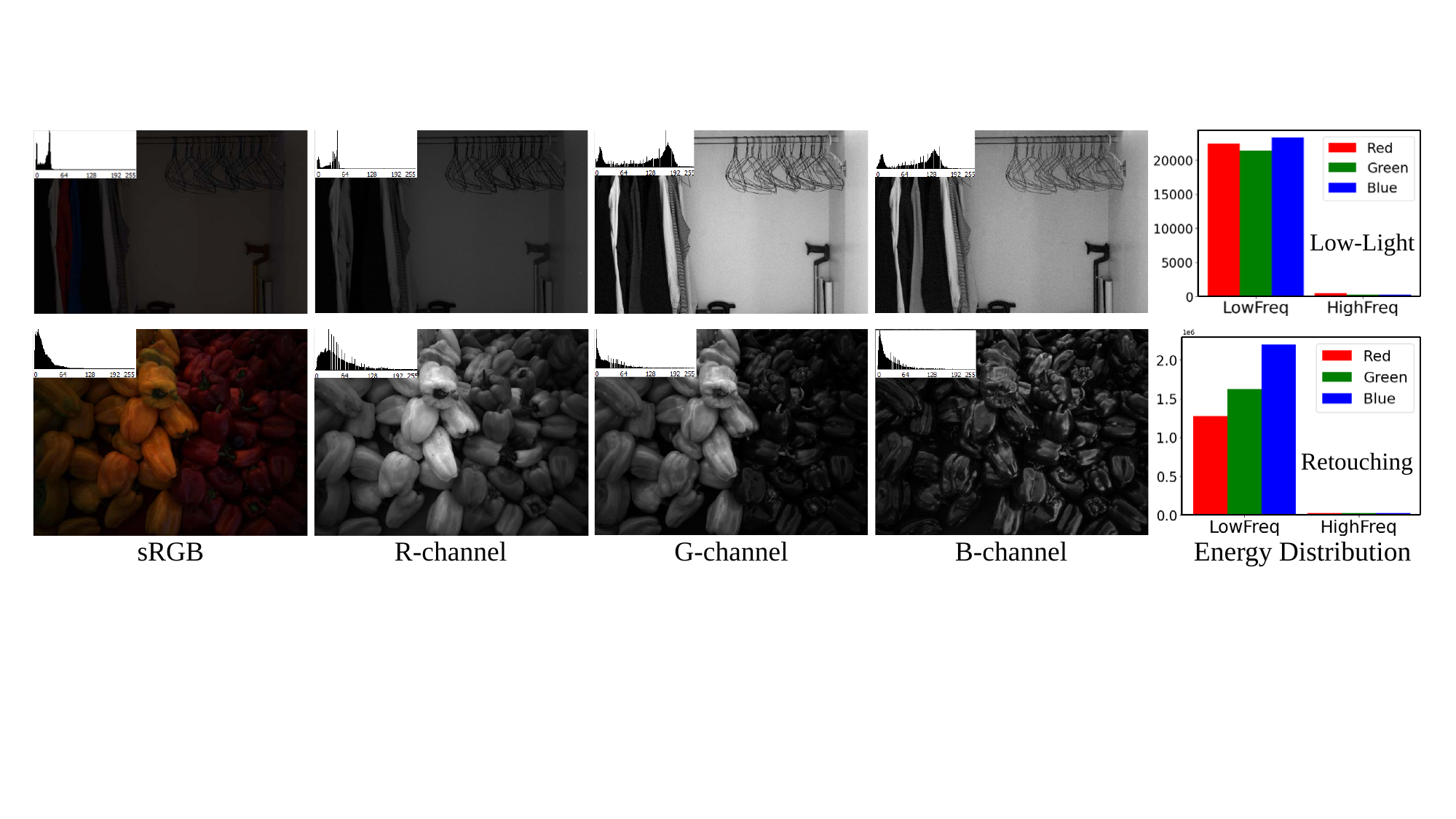}}  
    \vspace{-0.4cm}
    \caption {Motivation of LALNet. Different color channel differences and statistical DWT spectral energy distributions for different tasks.}
    \vspace{-0.3cm}
    \label{motivate}
\end{figure*}

Many tasks in computer vision aim to achieve light adaptation, including exposure correction~\citep{li2024cotf, huang2023learning}, image retouching~\citep{He_CSRnet_2020, Zhang_llfLUT_2023}, low-light enhancement~\citep{cai2023retinexformer, yi2023diff}, and tone mapping~\citep{cao2023unsupervised, Yang_SepLUT_2022}. 
The common goal of these light-related tasks is to adjust the light level of the scene to the perceptually optimal level, thereby revealing more visual details. 
However, due to the different characteristics of these light-related tasks, most of the current methods~\citep{Zeng_lut_2020, li2024cotf} are designed to deal with the above tasks individually and are difficult to apply to other light-related tasks.
For example, exposure correction~\citep{huang2022deep, zhang2019kindling} must adjust the brightness of both underexposed and overexposed scenes to achieve clearer images; image retouching~\citep{wang2023learning, su2024styleretoucher} aims to enhance the aesthetic visual quality of images affected by light defects, often requiring special attention to global light; low-light enhancement~\citep{wang2022uformer, liu2021retinex} reveals more details by boosting the brightness of dark areas, but requires special processing of noise; and tone mapping~\citep{Zhang_CLUT, wang2021real} preserves rich details by compressing high dynamic range light to low dynamic range, focusing more on adaptation to high dynamic range light.
The different characteristics of these tasks make existing methods inconsistent in performance on multiple tasks. 
Although some works~\citep{Yang_Cheng_2023} have attempted to perform light-related tasks with a unified architecture, the insufficient analysis of light-related task specificity has resulted in unsatisfactory performance compared to methods designed for these individual tasks.

\textit{Interestingly, can a framework be designed to handle these light-related tasks, just as the human visual system can adapt to a variety of lighting environments?} 
Motivated by this question, we aim to design a framework capable of handling multiple light enhancement tasks separately. 

To this end, we delve deep into analyzing the common light properties of these light-related tasks and utilize them to inspire the design of a multi-task framework.
We observe two key insights from light-related tasks: \textbf{i) different color channels have different light properties}; \textbf{ii) the channel differences reflected in the spatial and frequency domains are different.}
To analyze these differences, we employ the Discrete Wavelet Transform~\citep{shensa1992discrete} to decompose the input image into low-frequency and high-frequency components, and statistics on the energy distribution of the R/G/B channels based on the square of the pixel values separately.
Fig.~\ref{motivate} illustrates the color channel attributes of two light-related task images in the spatial and frequency domains.
It can be observed that the light properties of different channels differ significantly and that there is no fixed pattern between the different images. 
For example, for the first image, the G-channel exhibits a more balanced luminance distribution, while for the second image, the R-channel performs better in this regard. 
On the other hand, the frequency domain exhibits channel differences that are different from the spatial domain. For example, in the first image, the G-channel is brighter, but the G-channel does not have the highest energy distribution in the frequency domain.
This illustrates that capturing channel differences in the spatial and frequency domains is different. Channel differences cannot be fully characterized in the spatial or frequency domains alone.
%
%
Moreover, it is well known that the specific attributes~\citep{Yang_Cheng_2023, Zhang_llfLUT_2023} of light-related tasks are mainly embodied in the low-frequency components, whereas the details of the contents are more related to the high-frequency components. 
These findings highlight the importance of learning adaptive lighting by leveraging distinctive features of different color channels in the spatial and frequency domains.

Motivated by the above light properties, we propose a learnable adaptive lighting network, namely LALNet. 
Our method leverages the potential channel light differences to guide effective adaptive lighting. 
We decompose the light adaptation problem into two sub-tasks: (i) light adaptation, which addresses light variations under different light conditions, and (ii) detail enhancement, which preserves and refines image details while performing adaptive lighting. 
LALNet begins to learn adaptive light enhancement from the down-sampled version of the input image, optimizing for low computational complexity. 
To implement light adaptation, we propose a dual-branch architecture comprising channel separation and channel mixing. 
The channel separation branch employs the Dual Domain Channel Modulation module to extract color-separated features, focusing on light differences and color-specific luminance distributions for each channel in the spatial and frequency domains. 
In the channel mixing branch, we apply Mixed Channel Modulation and Light State Space Module to integrate color-mixed lighting information, capturing inter-channel relationships and lighting patterns that achieve harmonious light enhancement. 
A key component of our framework is Light Guided Attention (LGA), which utilizes color-separated features to guide color-mixed light information for adaptive lighting.
This mechanism enhances the network’s capability to perceive changes in channel luminance differences and ensure visual consistency and color balance across channels.
Consequently, our network is effectively adaptive to light variations while attending to feature differences across channels. 
Finally, we employ an iterative detail enhancement strategy to recover the image resolution level by level while enhancing the details. 
We conduct comprehensive experiments and demonstrate the state-of-the-art performance of our LALNet on four light-related tasks, as shown in Fig.~\ref{examples}.\\
Our contributions can be summarized as follows:
\vspace{-0.2cm}
\begin{itemize} 
\item We propose a multi-task light adaptation framework inspired by the common light property, namely the Learning Adaptive Lighting Network (LALNet).
\item We introduce the Dual Domain Channel Modulation to capture the light differences of different color channels and combine them with the traditional color-mixed features with Light Guided Attention.
\item Extensive experiments on four representative light-related tasks show that LALNet significantly outperforms state-of-the-art methods in benchmarking and that LALNet requires fewer computational resources.
\end{itemize}

\begin{figure*}[t]
    \centering	  
    \vspace{-0.2cm}
    \centering{\includegraphics[width=0.99\textwidth]{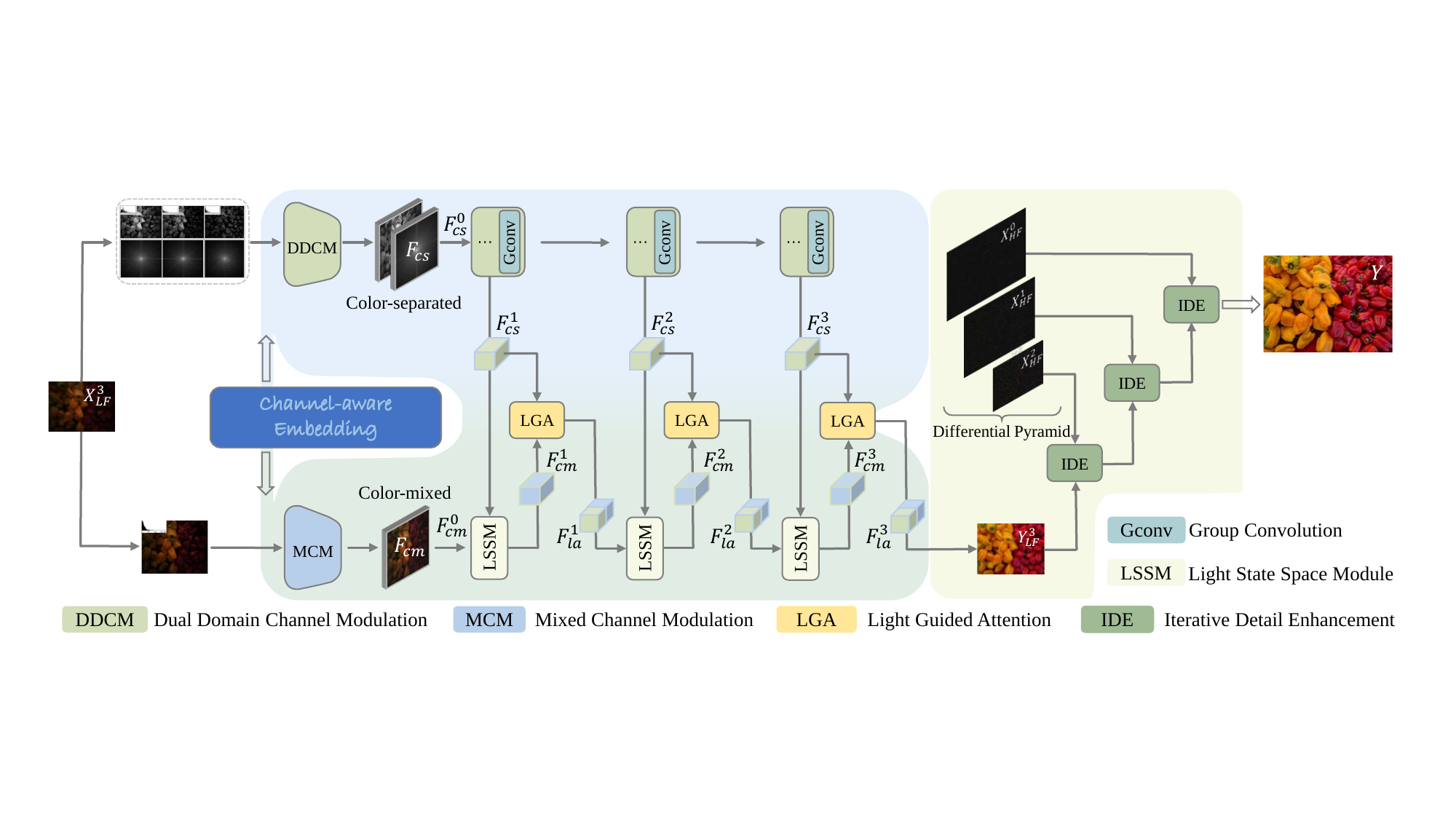}}  
    \vspace{-0.3cm}
    \caption {Architecture of LALNet for light adaptation. The core modules of LALNet are: (a) dual domain channel modulation (DDCM) that extracts color-separated features, focusing on light differences for each channel in the spatial and frequency domains, and (b) light guided attention (LGA) utilizes color-separated features to guide color-mixed light information for light adaptation.}
    \vspace{-0.25cm}
    \label{arch}
\end{figure*}

\vspace{-0.2cm}
\section{Related Work}
\vspace{-0.1cm}

\textbf{Exposure Correction.} 
Exposure correction aims to balance image brightness under varying lighting conditions~\citep{yang2020fidelity, nsampi2021learning, huang2022deep, li2024real}.
Early methods like RetinexNet~\citep{liu2021retinex} follow the Retinex theory to separately process illumination and reflectance. ZeroDCE~\citep{guo2020zero} estimates pixel-wise curves without reference images. However, these approaches mainly address underexposure and struggle with diverse real-world scenarios. LPNet~\citep{afifi2021learning} and FourierNet~\citep{huang2022deep} introduce multi-scale and frequency-aware designs for broader exposure handling. Recently, COTF~\citep{li2024real} proposed a collaborative framework for real-time correction, effectively integrating global and pixel-level adjustments.

\textbf{Image Retouching.} 
Image retouching focuses on restoring natural luminance and color distributions in a more perceptually faithful manner~\citep{moran2020deeplpf, liang2021ppr10k, gao2021real}. Some approaches reformulate the problem as curve estimation~\citep{kim2020global, li2020flexible}, while others like DeepLPF~\citep{moran2020deeplpf} optimize spatially adaptive filters for fine control.
Lookup table (LUT)-based models~\citep{Zeng_lut_2020, liang2021ppr10k} offer efficient inference by learning compact representations. CSRNet~\citep{He_CSRnet_2020} leverages conditional MLPs for adaptive enhancement, and GAN-based methods~\citep{chen2018deep, ni2020towards} enable unpaired learning, though often at the cost of interpretability and training stability.

\textbf{Tone Mapping.} 
Recent advancements in tone mapping have leveraged deep learning methods~\citep{Zhang_CLUT, Yang_SepLUT_2022, Zhang_llfLUT_2023, Hu_2022} to address the nonlinear mapping from HDR to LDR images. Hou et al. \cite{hou2017deep} applied CNNs to tone mapping tasks, establishing a foundation for subsequent research. Later works explored GANs for pixel-level accuracy~\citep{Cao_2020, Rana_DEEPTMO_2020, Panetta_2021}. Despite these advancements, issues such as halo artifacts and local inconsistencies persist. JointTM~\citep{Hu_2022} combined tone mapping and denoising using discrete cosine transforms, while HSVNet~\citep{zhang2019deep} leveraged HSV color space manipulation to reduce halos and enhance detail retention. Despite notable progress, existing methods often struggle to balance global and local tone mapping, resulting in unsatisfactory results in other tasks.

\textbf{Low-Light Image Enhancement.} 
Recent advances in low-light enhancement are largely driven by deep learning ~\citep{yang2023learning, liu2021swin, yang2025learning}. 
Models like DeepUPE~\citep{wang2019underexposed} and Retinexformer~\citep{cai2023retinexformer} build on Retinex theory for illumination decomposition. Hybrid architectures such as SNRNet~\citep{xu2022snr} and Restormer~\citep{zamir2022restormer} incorporate Transformer designs for long-range dependencies. RetinexMamba~\citep{bai2024retinexmamba} introduces the State Space Model to improve efficiency. However, Retinex-based methods~\citep{cai2023retinexformer, liu2021retinex, bai2024retinexmamba} are based on the theory of separated illumination and reflection, but they usually assume smooth and uniform lighting conditions, which may not hold in realistic scenes involving complex lighting variations. Moreover, these methods typically work in luminance or reflection space, where high-frequency details may be distorted during decomposition. 

\vspace{-0.1cm}
\section{Methods}
\label{method}
\vspace{-0.1cm}
\subsection{Motivation}

Previous studies~\citep{cai2023retinexformer, li2024cotf, Zhang_llfLUT_2023, su2024styleretoucher} for light-related tasks, such as tone mapping and low-light enhancement, are often tailored to individual tasks, leading to suboptimal performance across multiple scenarios. These frameworks typically fail to account for the common properties shared across different lighting-related tasks, which limits their generalizability. As a result, many frameworks are either overly specialized or inefficient when faced with multiple tasks. This leads to performance inconsistencies, especially when frameworks designed for specific tasks are applied to others. For instance, Retinexformer focuses on separating reflection and illumination to enhance low-light images, but its underlying Retinex theory is inapplicable to tasks such as tone mapping and image retouching. This limitation is evident in scenarios where low-light enhancement methods struggle to maintain color fidelity during tone mapping.

Our motivation is rooted in the observation that, despite the diverse nature of light-related tasks, there are key shared properties: \textbf{distinct light properties across color channels} and \textbf{channel differences in spatial and frequency domains.} 
These channel differences manifest differently in both the spatial and frequency domains, further complicating the task of adaptive lighting.
To address these issues, we aim to design a multi-task framework that adapts to different lighting conditions more effectively than previous frameworks that focus on individual tasks. By analyzing these shared light properties across multiple tasks, our framework seeks to capture the subtle differences between color channels and ensure consistent and balanced visual outcomes across various lighting conditions.

\subsection{Framework Overview}

The overall pipeline of LALNet is illustrated in Fig.~\ref{arch}. 
LALNet is composed of two components: light adaptation and detail enhancement. Given a low-quality (LQ) input image $\mathbf{X}^0$, our goal is to generate a high-quality (HQ) output $\mathbf{Y}$ with optimal light. 
We begin to learn adaptive light from the down-sampled version of the input image $\mathbf{X}^{3}_\textrm{LF}$, optimizing for low computational complexity.
Subsequently, we employ the two-branch structure for extracting light features, containing color separation and color mixing branches.
The channel separation branch employs the DDCM and group convolution to extract color-separated feature $\mathbf{F}_\textrm{cs}$, focusing on light differences and color-specific luminance distributions for each channel in the spatial and frequency domains.
In the channel mixing branch, we utilize mixed channel modulation (MCM) combined with the light state space module (LSSM) to extract color-mixed feature $\mathbf{F}_\textrm{cm}$, promoting cross-channel interaction and achieving balanced light enhancement. 
This can be expressed mathematically as:
\begin{equation} 
\mathbf{F}_\mathrm{cs} = \mathrm{GConv}(\mathrm{DDCM}(\mathbf{X}^3_{LF})),
\end{equation}
\begin{equation} 
\mathbf{F}_\mathrm{cm} = \mathrm{LSSM}(\mathrm{MCM}(\mathbf{X}^3_{LF}), \mathbf{F}_\mathrm{cs}).
\end{equation}
To emphasize the light differences in different channels, we introduce Light Guided Attention, which injects the color-separated features into color-mixed features to obtain the light adaptive feature $F_{la}$, which is described as:
\begin{equation} 
\mathbf{F}_\mathrm{la} = \mathrm{LGA}(\mathbf{F}_\mathrm{cm}, \mathbf{F}_\mathrm{cs}).
\end{equation}
This process ensures consistent and uniform light adaptation across the entire image and eliminates color distortion caused by channel crosstalk. 
Finally, we integrate the low- and high-frequency via learnable differential pyramid \cite{yang2024learning} and iterative detail enhancement, progressively refining image resolution and enhancing fine details.

\vspace{-0.1cm}
\subsection{Light Adaptation}

In the literature, we generally utilize the traditional convolutions to convolve with all channels for light-related tasks, generating RGB-mixed features. 
This operator can capture the interaction information and shared features among channels. 
However, this also amplifies the luminance non-uniformity and noise existing in the three channels. 
Notably, for light-related tasks, we have observed that characteristic differences between the RGB channels and the spatial and frequency domains exhibit different differences. There is also no consistent pattern across images.
As shown in Fig.~\ref{motivate}, the three channels exhibit distinct differences in luminance, with one channel usually being closer to the ground truth. If we only utilize color-mixed features to adapt to light, the negative interference between channels will also spread to all channels. 
Therefore, we introduce an additional branch that extracts channel-separated features alongside the channel-mixed features. Channel-mixed features are responsible for capturing mixed luminance and color information, while channel-separated features guide the network to focus on channel differences. 
This design prompts the network to adapt to light while attending to differences across channels.

\vspace{-0.1cm}
\subsubsection {Color Separation Representation}

Based on the analysis in Sec.~\ref{intro}, the spatial and frequency domains reflect different channel differences. Therefore, we design DDCM to capture the color-separated features.

\textbf{Dual Domain Channel Modulation.}
To avoid cross-channel interference between operating channels, we process each channel independently in the frequency and spatial domains and introduce learnable parameters to modulate the channels.
After frequency domain processing, the images are inverted back to the spatial domain. 
Then, to complement the color-separated feature representation, we utilize channel attention to capture the color-separated features in the spatial domain.
Specifically, given an input image $\mathbf{X}$, each channel of the image is denoted as $\mathbf{X}_i$ ($i =1,2,3$). We perform a 2D fast Fourier Transform (FFT) for $\mathbf{X}_i$ to obtain the frequency domain representation:
\begin{equation}
\mathbf{S}_i(u,v) = \mathcal{F}(\mathbf{X}_i)(u,v) = \mathrm{FFT2}(\mathbf{X}_i),
\vspace{-0.1cm}
\end{equation}
where $\mathbf{S}_i(u,v)=\mathbf{R}_i(u,v)+j\cdot \mathbf{I}_i(u,v)$, $\mathbf{R}_i(u,v)$ and $\mathbf{I}_i(u,v)$ denote the real and imaginary parts, respectively. Then, we perform convolution operations on the $\mathbf{R}_i(u,v)$ and $\mathbf{I}_i(u,v)$, respectively:
\vspace{-0.1cm}
\begin{equation}
\mathbf{\hat{R}}_i(u,v) = \mathbf{W}_{R_i} * \mathbf{R}_i(u,v),    \quad 
\mathbf{\hat{I}}_i(u,v) = \mathbf{W}_{I_i} * \mathbf{I}_i(u,v),
\end{equation}
where $\mathbf{W}_{R_i}$ and $\mathbf{W}_{I_i}$ are the convolution kernels, $*$ denote convolution operation. 
Subsequently, we reorganize the decoupled real and imaginary parts into frequency-domain signals, and perform the Inverse Fourier Transform to obtain the decoupled time-domain information as follows:
\begin{equation} 
\mathbf{S}'_i(u,v)=\mathbf{R}'_i(u,v)+j\cdot \mathbf{I}'_i(u,v),
\end{equation}
\begin{equation} 
\mathbf{X}'_i = \mathcal{F}^{-1}(\mathbf{S}'_i(u,v)) = \mathrm{IFFT2}(\mathbf{S}'_i).
\end{equation}
Finally, after concatenating channels, we capture the separated features of the image in the spatial domain through the channel attention module to further enhance the color-separated feature representation.
\begin{equation} 
\mathbf{F}_\mathrm{cs} = \mathrm{CAB}(\mathrm{Concat}(\mathbf{X}'_1,\mathbf{X}'_2,\mathbf{X}'_3)).
\end{equation}

\vspace{-0.2cm}
\subsubsection {Color Mixing Representation}

In parallel, we introduce mixed channel modulation for extracting channel-mixed features. 
Since light patterns often exhibit global characteristics~\citep{Rieke_Rudd_2009, Yang_Cheng_2023}, inspired by~\citep{finder2024wavelet}, we employ wavelet transform to achieve channel-mixed features $\textrm{F}_\textrm{cm}$. 
The process begins with the extraction of small-scale features using a small convolutional kernel to capture local information. These features are then passed through a wavelet transform (WT), where the generated large-scale features modulate the small-scale features, enabling the network to better integrate global light representation. The process can be represented as follows:
\begin{equation} 
\mathbf{cA}, \mathbf{cH}, \mathbf{cV}, \mathbf{cD} = \mathrm{WT}(\mathrm{Conv}_{3 \times 3}(\mathbf{X})),
\end{equation}
where $\mathbf{cA}, \mathbf{cH}, \mathbf{cV}, \mathbf{cD}$ represent the components of the 2D wavelet transform.
Afterward, the modulated features are concatenated and further passed the convolutional layer.
\begin{equation} 
\mathbf{F}^0_\mathrm{cm} = \mathrm{Conv}_{3 \times 3}(\mathrm{Concat}(\mathbf{cA}, \mathbf{cH}, \mathbf{cV}, \mathbf{cD})).
\end{equation}
To enhance the network's capability to capture global light information, we introduce the Light State Space Module (LSSM), which supplements mixed-channel modulation. The LSSM is designed to efficiently capture long-range dependencies with lower computational overhead than transformer-based methods.
For feature integration and expansion, LSSM begins by integrating channel-mixed features with channel-separated features. This integrated feature is then expanded to a dimensionality of $2C$ via a linear layer. Following this expansion, the feature is divided into two distinct components, $\mathbf{F}_1$ and $\mathbf{F}_2$, according to the channel dimensions. Therefore, the channel-separated feature $\mathbf{F}^1_\mathrm{cs}$, along with the newly formed $\mathbf{F}_1$ and $\mathbf{F}_2$, serve as inputs to three parallel processing streams. First Stream: Feature $\mathbf{F}_1$ undergoes an initial expansion to $\eta C$ channels through a linear layer, followed by depth-wise convolution, SiLU, 2D selective scanning (SS2D) \cite{guo2024mambair}, and LayerNorm. This sequence refines the representation of $\mathbf{F}_1$, emphasizing its spatial and channel-wise characteristics. Second Stream: Feature $\mathbf{F}^1_\mathrm{cs}$ is processed directly using SS2D, capturing comprehensive global context without additional transformations. Third Stream: Feature $\mathbf{F}_2$ is subjected only to SiLU, preserving its original characteristics while enabling non-linear transformations that enrich its representation.
Subsequently, the global information extracted from the first two streams is fused. This fused information is then multiplied with the output of the third stream. By doing so, the LSSM effectively integrates detailed local and global light patterns, enhancing the overall sensitivity of the network to varying lighting conditions.
The whole process can be represented as follows:
\begin{equation} 
\mathbf{F}_1, \mathbf{F}_2 = \mathrm{Chunk}(\mathrm{Linear}(\mathbf{F}^0_\mathrm{cm} + \mathbf{F}^1_\mathrm{cs})),
\end{equation}
\begin{equation} 
\mathbf{F}'_{1} = \mathrm{SS2D}(\mathrm{SiLU}(\mathrm{DWConv}(\mathbf{F}_1))) + \mathrm{SS2D}(\mathbf{F}^1_\mathrm{cs})
\end{equation}
\begin{equation} 
\mathbf{F}'_2 = \mathrm{SiLU}(\mathbf{F}_2), \quad
\mathbf{F}^1_\mathrm{cm} = \mathrm{MLP}(\mathrm{LN}((\mathbf{F}'_1 \otimes \mathbf{F}'_2))),
\end{equation}
where $\textrm{Linear}(\cdot)$ denote linear projection, $\otimes$ denotes the Hadamard product.

\subsubsection{Light Guided Attention}

Although LSSM performs well in capturing long-range dependencies, it still faces problems such as local information forgetting and channel redundancy. Moreover, color mixed features ignore the feature differences between different channels, treating them equally in the network.
However, in light-related tasks, we have observed significant differences between color channels, with no consistent pattern across images. These differences are crucial for adaptive lighting. 
For this reason, we propose to inject color-separated features into color-mixed features by light guided attention to perceive channel differences.

Specifically, for the first LGA module, we input the channel-mixed features $\mathbf{F}^1_\textrm{cm}$ from LSSM and the channel-separated features $\mathbf{F}^1_\textrm{cs}$ from group convolution into the LGA. 
Subsequently, the input $\mathbf{F}^1_\textrm{cm}$ is processed through a $1\times1$ convolution followed by a depthwise convolution, producing $\mathbf{K}$ and $\mathbf{V}$ tensors with doubled the number of channels. This can be expressed mathematically as:
\begin{equation} 
\mathbf{K}, \mathbf{V} = \mathrm{Conv}_{3 \times 3}(\mathrm{Conv}_{1 \times 1}(\mathbf{F}^1_\mathrm{cm})).
\end{equation}
The query $\mathbf{Q}$ is then generated from the channel-separated features $\mathbf{F}^1_\mathrm{cs}$:
\begin{equation} 
\mathbf{Q} = \mathrm{Conv}_{3 \times 3}(\mathrm{Conv}_{1 \times 1}(\mathrm{GConv}_{3 \times 3}(\mathbf{F}^1_\mathrm{cs}))).
\end{equation}
We compute the attention weights by the dot product between $\mathbf{Q}$ and $\mathbf{K}$, normalized by the softmax function, and multiplied by $\mathbf{V}$ to obtain the updated features:
\begin{equation} 
\mathrm{Attention}(\mathbf{Q}, \mathbf{K}, \mathbf{V}) = \mathrm{softmax}(\frac{\mathbf{QK}^T}{\sqrt{d_K}} \times \tau)\mathbf{V},
\end{equation}
where $d_K$ is the dimension of $\mathbf{K}$ and $\tau$ denotes the scaling factor. 
It can be remarked that we utilize channel-separated features as $\mathbf{Q}$ vectors to motivate the model to focus on channel differences. 
In summary, the design of LGA enhances the adaptive representation of image features in both spatial and channel dimensions and improves the network's ability to capture dependencies between image channels.
After LGA processing, we can obtain the low-resolution 
light-adaption output $\mathbf{Y}^3_\textrm{LF}$. 
Subsequently, we utilize the iterative detail enhancement strategy 
to enhance the detail of $\mathbf{Y}^3_\textrm{LF}$, which is introduced in the following.

\begin{table*}[ht]
    \centering
    \vspace{-0.2cm}
    \caption{Quantitative results of exposure correction methods on the SCIE dataset. "/" denotes the unavailable source code. Metrics with $\uparrow$ and $\downarrow$ denote higher better and lower better. The best and second results are in red and blue, respectively.}
    \label{tab:EC}
    \scalebox{0.78}{
    \begin{tabular}{c|cc|cc|ccccc}
        \toprule[1pt]
        \multirow{3}{*}{Method} & \multicolumn{9}{c}{Exposure Correction in SCIE}  \\
    \cline{2-10}
    & \multicolumn{2}{c|}{Under} & \multicolumn{2}{c|}{Over} & \multicolumn{5}{c}{Average} \\
     &  PSNR$\uparrow$ & SSIM$\uparrow$  &  PSNR$\uparrow$ & SSIM$\uparrow$ &  PSNR$\uparrow$ & SSIM$\uparrow$ & LPIPS$\downarrow $   & NIQE$\downarrow$ & MUSIQ$\uparrow$\\ 
    \midrule
    URtinexNet ~\citep{wu2022uretinex}    & 17.39 & 0.6448 & 7.40 & 0.4543  & 12.40 & 0.5496 & 0.3549  & 12.78 & 49.11 \\
    DRBN ~\citep{yang2020fidelity}    & 17.96 & 0.6767 & 17.33 & 0.6828 & 17.65 & 0.6798 & 0.3891 & 12.06 & 48.77\\
    SID ~\citep{chen2018deep}      & 19.51 & 0.6635 & 16.79 & 0.6444  & 18.15 & 0.6540 & 0.2417 & 11.79 & 51.07 \\
    CSRNet ~\citep{He_CSRnet_2020}  & 21.43 & 0.6789 & 20.13 & 0.7250  & 20.78 & 0.7019 & {\color{blue}\underline{0.1390}}  & 10.59 & {\color{blue}\underline{61.79}}\\
    MSEC ~\citep{afifi2021learning}     & 19.62 & 0.6512 & 17.59 & 0.6560  & 18.58 & 0.6536 & 0.2814  & /  & / \\
    SID-ENC ~\citep{huang2022exposure} & 21.30 & 0.6645 & 19.63 & 0.6941 & 20.47 & 0.6793 & 0.2797 & 11.49 & 52.29 \\
    DRBN-ENC ~\citep{huang2022exposure} & 21.89 & 0.7071 & 19.09 & 0.7229 & 20.49 & 0.7150 & 0.2318 & 11.23 &54.15 \\
    CLIP-LIT ~\citep{liang2023iterative} & 15.13 & 0.5847 & 7.52 & 0.4383 & 11.33 & 0.5115 & 0.3560  & /  & / \\
    FECNet ~\citep{huang2022deep}          & 22.01 & 0.6737 & 19.91 & 0.6961  & 20.96 & 0.6849 & 0.2656  &11.05  &53.73 \\
    FECNet+ERL ~\citep{huang2023learning}   & 22.35 & 0.6671 & 20.10 & 0.6891  & 21.22 & 0.6781 & /  & /  & / \\  
    Retinexformer ~\citep{cai2023retinexformer}   & 23.75  &0.7157  & 22.13  & 0.7466 & 22.94  & 0.7310  & 0.1714    & 10.37 & 55.67\\
    CoTF ~\citep{li2024cotf}     & 22.90 & 0.7029 & 20.13 & 0.7274  & 21.51 & 0.7151 & 0.1924 & 10.19  & 51.61\\
    RetinexMamba ~\citep{bai2024retinexmamba}   & 23.56  & {\color{blue}\underline{0.7212}}  & 21.59  & 0.7384 &22.58   &  0.7298  & 0.1856   & 10.35 & 53.67 \\
    \midrule 
    LALNet-Tiny  & {\color{blue}\underline{23.86}} & 0.7197 & {\color{blue}\underline{22.26}} & \textbf{{\color{red}0.7510}} & {\color{blue}\underline{23.06}} & {\color{blue}\underline{0.7354}} & \textbf{{\color{red}0.1280}} & \textbf{{\color{red}8.93}} & \textbf{{\color{red}63.01}} \\
    LALNet   &  \textbf{{\color{red}24.63}}	&  \textbf{{\color{red}0.7270}}	& \textbf{{\color{red}22.95}}	& {\color{blue}\underline{0.7473}}	& \textbf{{\color{red}23.80}}	& \textbf{{\color{red}0.7372}} & 0.1397   & {\color{blue}\underline{9.34}} & 61.49 \\
    \bottomrule[1pt] 
    \end{tabular}
    }
    \vspace{-0.25cm}
\end{table*}

\subsection{Detail Enhancement}
To achieve faithful reconstruction, we apply a learnable differential pyramid (LDP) \cite{yang2024learning} to capture high-frequency details. Through LDP, we obtain the multi-scale high-frequency features $\mathbb{X}_\textrm{HF}=\{\mathbf{X}^{0}_\textrm{HF}, \ldots, \mathbf{X}^{L-1}_\textrm{HF}\}$, tapering resolutions from $H \times W$ to $\frac{H}{2^{L-1}} \times \frac{W}{2^{L-1}}$. $L$ denotes the number of pyramid levels ($L$=3 in our framework). 
Using the high-frequency information $\mathbb{X}_\textrm{HF}$ captured, we employ an iterative detail enhancement to progressively refine the light-adaption image $\mathbf{Y}^L_\textrm{LF}$. 
Specifically, for the $l_{th}$ pyramid, we first up-sample the low-frequency image $\mathbf{Y}^l_\textrm{LF}$ and concatenate it with the HF component $\mathbf{X}^{l-1}_\textrm{HF}$, then feed it into a residual network to predict a refinement mask $\mathbf{M}^{l-1}$. This mask allows pixel-by-pixel refinement of the HF component, which is subsequently added to the up-sampling $\mathbf{Y}^l_\textrm{LF}$ to generate the reconstructed result of the current layer $\mathbf{Y}^{l-1}_{\textrm{LF}}$. The process at the $l_{th}$ pyramid is formulated as:
\begin{equation}
\mathbf{M}^{l-1} = \mathrm{Res}(\mathrm{Concat}(\mathrm{Up}(\mathbf{Y}^{l}_{\mathrm{LF}}), \mathbf{X}^{l-1}_{\mathrm{HF}})), 
\end{equation}
\begin{equation}
\mathbf{Y}^{l-1}_{\mathrm{LF}} = \mathrm{Up}(\mathbf{Y}^{l}_{\mathrm{LF}}) + (\mathbf{X}^{l-1}_{\mathrm{HF}} \mathbf{M}^{l-1}),
\end{equation}
where $\textrm{Res}(\cdot)$ and $\textrm{Up}(\cdot)$ denote the residual block and up-sampling, respectively. 

\begin{figure*}[htb]
    \centering
    \includegraphics[width=0.99\textwidth]{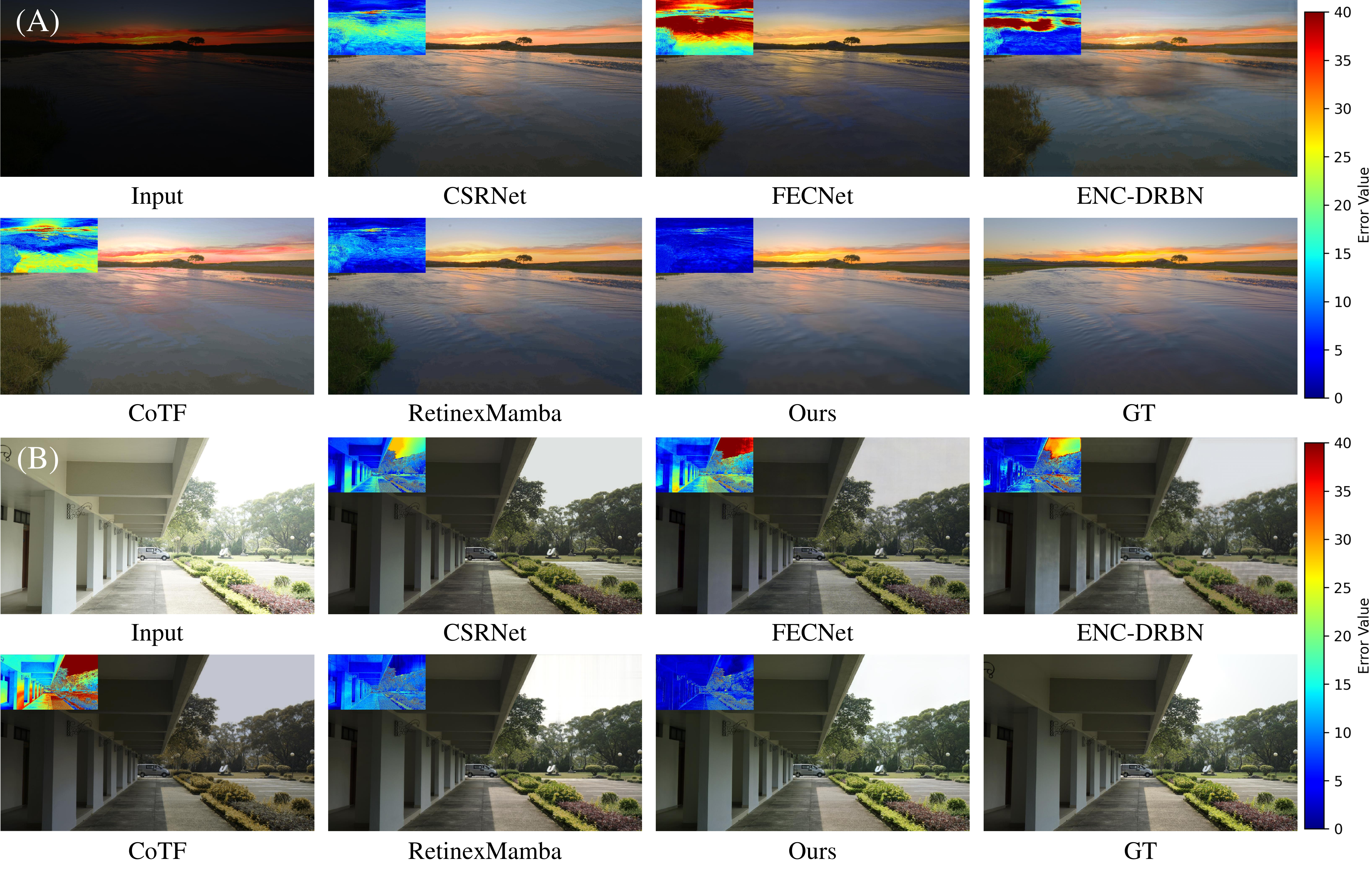}
    \vspace{-0.4cm}
    \caption{Visual comparisons between our LALet and the SOTA methods on the SCIE dataset. (Zoom in for best view.) The error maps in the upper left corner facilitate a more precise determination of performance differences.}
    \vspace{-0.2cm}
\label{scie}
\end{figure*}
\begin{table*}[t]
\centering
\caption{Quantitative results of image retouching methods. "/" denotes the unavailable source code.}
\label{tab:IR_TM}
\scalebox{0.91}{
\small
\begin{tabular}{c|c|ccccccc}
\toprule[1pt]
    \multirow{2}{*}{Method} & \multirow{2}{*}{\#Params}  &\multicolumn{7}{c}{Image Retouching in HDRPlus} \\
    \cline{3-9}
     &  &  PSNR$\uparrow$ & SSIM$\uparrow$  & TMQI$\uparrow$  & LPIPS$\downarrow $ & $\bigtriangleup$E$\downarrow $ & NIQE$\downarrow$ & MUSIQ$\uparrow$\\
    \midrule
    HDRNet ~\citep{Gharbi_hdrnet_2017} &482K & 24.15 & 0.845 & 0.877 & 0.110 & 7.15 & 10.47 & 68.73\\
    CSRNet ~\citep{He_CSRnet_2020}     &37K & 23.72 & 0.864 & 0.884 & 0.104 & 6.67  & 10.99 & 67.82\\
    DeepLPF ~\citep{moran2020deeplpf}  &1.72M & 25.73 & 0.902 & 0.877 & 0.073 & 6.05  & 10.35 & 70.02\\
    LUT ~\citep{Zeng_lut_2020}         &592K & 23.29 & 0.855 & 0.882 & 0.117 & 7.16  & 11.36  & 67.67 \\
    CLUT ~\citep{Zhang_CLUT}           &952K & 26.05 & 0.892 & 0.886 & 0.088 & 5.57 & 11.19 & 67.39\\
    sLUT ~\citep{wang2021real}       &4.52M & 26.13 & 0.901 &  /    & 0.069 & 5.34  & / &/ \\ 
    SepLUT ~\citep{Yang_SepLUT_2022}   &120K & 22.71 & 0.833 & 0.879 & 0.093 & 8.62 & 12.26 & 67.89\\
    Restormer ~\citep{zamir2022restormer} & 26.1M &  25.93	& 0.900	& 0.883	& 0.050	& 6.59  &10.49 & 68.92\\
    LLFLUT ~\citep{Zhang_llfLUT_2023} &731K & 26.62 & 0.907 &  /    & 0.063 & 5.31  & / &/ \\  
    CoTF ~\citep{li2024cotf}          &310K & 23.78 & 0.882 & 0.876 & 0.072 & 7.76  & 11.54 &68.07 \\
    Retinexformer ~\citep{cai2023retinexformer} & 1.61M & 26.20 & 0.910 & 0.879 & 0.046 & 6.14 & 10.75 &68.93 \\
    RetinexMamba ~\citep{bai2024retinexmamba} & 4.59M & 26.81	& 0.911	& 0.880	& 0.047	& 5.89 & 10.52 &69.02 \\  
    MambaIR ~\citep{guo2024mambair} & 4.31M & 28.09	& {\color{blue}\underline{0.943}}	& 0.879	& {\color{blue}\underline{0.028}}	& 5.31 & 10.76 & {\color{blue}\underline{70.05}} \\  
    \midrule 
    LALNet-Tiny &230K & {\color{blue}\underline{29.30}} & 0.939 & {\color{blue}\underline{0.886}} & 0.030 & {\color{blue}\underline{5.04}}  & {\color{blue}\underline{9.70}} & 69.98\\
    LALNet  &2.45M &\textbf{{\color{red}30.24}}	& \textbf{{\color{red}0.944}}	& \textbf{{\color{red}0.888}}	& \textbf{{\color{red}0.027}}	& \textbf{{\color{red}4.52}} & \textbf{{\color{red}9.82}} & \textbf{{\color{red}70.25}} \\
    \bottomrule[1pt]
\end{tabular}}
\vspace{-0.25cm}
\end{table*}

\vspace{-0.2cm}
\section{Experiments}
\vspace{-0.1cm}
\subsection{Experimental settings}

\textbf{Datasets.} We evaluate our method on four representative light-related tasks: exposure correction (SCIE~\citep{cai2018learning}), image retouching (HDR+ Burst Photography~\citep{Hasinoff_hdrplus_2016}), low-light enhancement (LOL dataset~\citep{wei2018deep}), and tone mapping (HDRI Haven~\citep{yang2024learning}. Following the settings of~\citep{huang2022exposure} for SICE, it contains 1000 training images and 24 test images. The HDR+ dataset is a staple for image retouching, especially in mobile photography. We utilize 675 image sets for training and 248 for testing. The LOL dataset~\citep{wei2018deep} contains 500 image pairs in total, with 485 pairs used for training and 15 test images. The HDRI Haven dataset is a new benchmark for evaluating tone mapping~\citep{su2021explorable, cao2023unsupervised}, which includes 570 HDR images of diverse scenes under various light conditions. We select 456 image sets for training and 114 for testing. 

\textbf{Implementation details.} We implement our model with Pytorch on the NVIDIA L40s GPU platform. The model is trained with the Adam optimizer ($\beta_{1}=0.9$, $\beta_{2}=0.999$) for $4\times 10^{5}$ iterations. The learning rate is initially set to $1\times 10^{-4}$. We adopt traditional PSNR and SSIM metrics on the RGB channel to evaluate the reconstruction accuracy. We also employ TMQI~\citep{Yeganeh_Wang_2013}, LPIPS~\citep{zhang2018unreasonable}, and CIELAB color space~\citep{zhang1996spatial} to evaluate image quality and perceptual quality, respectively.

\subsection{Comparison with State-of-the-Arts} 

\textbf{Quantitative comparison.} The performance of the proposed multi-task framework is evaluated on four light-related image enhancement tasks, namely, (1) exposure correction, (2) image retouching, (3) low-light enhancement, and (4) tone mapping. We quantitatively compare the proposed method with a wide range of state-of-the-art light-related methods in Tab. \ref{tab:EC}, Tab. \ref{tab:IR_TM}, Tab. \ref{tab:TM}, and Tab. \ref{tab:LLE}. 
For exposure correction, as shown in Tab. \ref{tab:EC}, our method improves \textbf{2.29 dB} PSNR and 0.0221 SSIM compared to the CoTF~\citep{li2024cotf} (CVPR24) method. 
For image retouching, as shown in Tab. \ref{tab:IR_TM}, the proposed LALNet outperforms all the previous SOTA methods by a large margin. Specifically, our method significantly outperforms the SOTA methods MambaIR~\citep{guo2024mambair}, RetinexFormer~\citep{cai2023retinexformer}, LLFLUT~\citep{Zhang_llfLUT_2023} and CoTF~\citep{li2024cotf}, RetinexMamba~\citep{bai2024retinexmamba}, improving PSNR by \textbf{2.15 dB} in the HDR+ dataset. 
LALNet-Tiny is a lightweight variant of LALNet (fewer feature channels and fewer LSSM blocks.) 
LALNet-Tiny has only $230K$ parameters and $1.75$ GFLOPs, but the performance is also significantly better than other SOTA methods. 
For low-light enhancement, our LALNet significantly outperforms SOTA methods on the LOL-v1 dataset while requiring moderate computational and memory costs. Compared to the best recent method RetinexMamba, LALNet improves PSNR by \textbf{1.23 dB} and SSIM by 0.027, and LALNet only costs \textbf{16\%} (6.70 / 42.82) GFLOPs.
For tone mapping, Tab. \ref{tab:TM} reports the quantitative results on the HDRI Haven dataset. We can see that our method has the best overall performance.   
\begin{table}[t]
\centering
\vspace{-0.2cm}
\caption{Quantitative results of tone mapping methods. "/" denotes the unavailable source code.}
\label{tab:TM}
\small
\scalebox{0.73}{
\begin{tabular}{c|ccccc}
\toprule[1pt]
    \multirow{2}{*}{Method}  &\multicolumn{5}{c}{Tone Mapping in HDRI Haven} \\
    \cline{2-6}
     &  PSNR$\uparrow$ & SSIM$\uparrow$  & TMQI$\uparrow$  & LPIPS$\downarrow $ & $\bigtriangleup$E$\downarrow $   \\
    \midrule
    UPE ~\citep{Wang_UPE_2019}         &  23.58 & 0.821  & 0.917  & 0.191 & 10.85            \\
    HDRNet ~\citep{Gharbi_hdrnet_2017}  & 25.33 & 0.912 & 0.941 & 0.113 & 7.03             \\
    CSRNet ~\citep{He_CSRnet_2020}      & 25.78 & 0.872 & 0.928 & 0.153 & 6.09            \\
    DeepLPF ~\citep{moran2020deeplpf}   & 24.86 & 0.939 & 0.948 & 0.077 & 7.64           \\
    LUT ~\citep{Zeng_lut_2020}          & 24.52 & 0.846 & 0.912 & 0.171 & 7.33         \\
    CLUT ~\citep{Zhang_CLUT}           & 24.29 & 0.836 & 0.908 & 0.169 & 7.08  \\
    LPTN ~\citep{Liang_lptn_2021}      & 26.21 & 0.941 & 0.954 & 0.113 & 8.82 \\
    SepLUT ~\citep{Yang_SepLUT_2022}   & 24.12 & 0.854 & 0.915 & 0.165 & 8.03        \\
    Restormer ~\citep{zamir2022restormer}  & 27.30	& 0.954	& 0.948	& 0.032	& 5.67 \\ 
    CoTF  ~\citep{li2024cotf}          & 26.65 & 0.935 & 0.948 & 0.098 & 5.84  \\
    Retinexformer ~\citep{cai2023retinexformer} & 27.73	& 0.955	& 0.949	& 0.030 & 5.41 \\
    RetinexMamba ~\citep{bai2024retinexmamba}  & 28.60	& 0.955	& 0.953	& 0.032	& 5.12 \\  
    \midrule 
    LALNet-Tiny  & {\color{blue}\underline{31.17}} & {\color{blue}\underline{0.962}} & {\color{blue}\underline{0.959}} & {\color{blue}\underline{0.026}} & {\color{blue}\underline{4.23}}\\
    LALNet  & \textbf{{\color{red}32.46}}  & \textbf{{\color{red}0.969}}	& \textbf{{\color{red}0.961}}	& \textbf{{\color{red}0.019}}	& \textbf{{\color{red}3.58}} \\
    \bottomrule[1pt]
\end{tabular}}
\vspace{-0.5cm}
\end{table}

\begin{figure*}[t]
    \centering
    \includegraphics[width=0.99\textwidth]{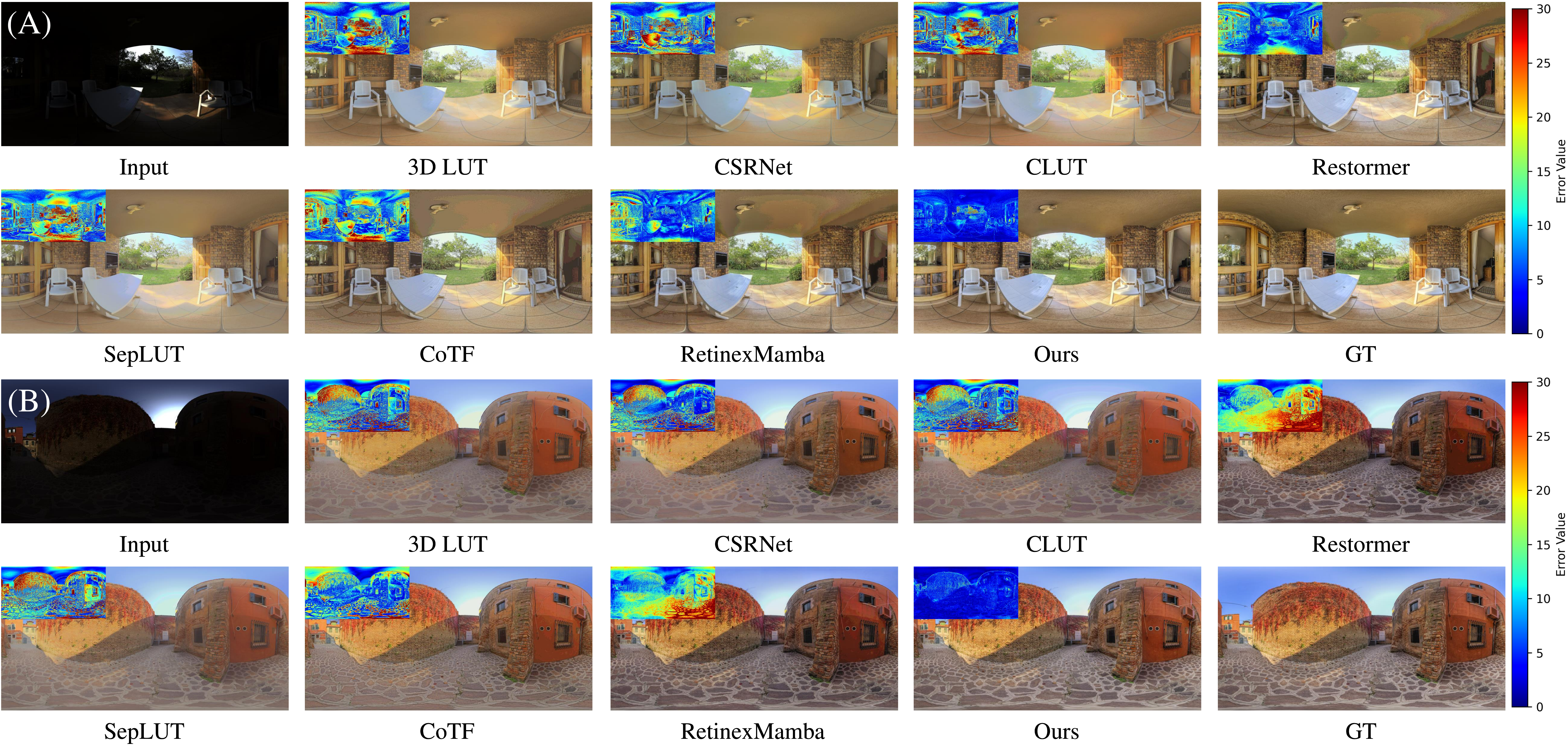}
    \vspace{-0.3cm}
    \caption{Visual comparisons between our LALet and the state-of-the-art methods on the HDRI Haven dataset (Zoom-in for best view). The error maps in the upper left corner facilitate a more precise determination of performance differences.}
    \vspace{-0.3cm}
\label{haven}
\end{figure*}

\textbf{Qualitative results.} Visual comparison of LALNet and state-of-the-art light-related image enhancement methods are shown in Fig. \ref{scie}, Fig. \ref{haven}, Fig. \ref{hdrplus}, and Fig~\ref{lol}. Please zoom in for better visualization. To better visualize the performance differences of various methods, we present an error map to show the differences between the results of each method and the target image, as shown in the upper left corner of the image. In the error map, the red area indicates a larger difference, while the blue area indicates that the two are closer. Notably, error maps have no special units and only indicate errors. These figures illustrate that our LALNet consistently delivers visually appealing results on light-related tasks. Results reveal that the proposed method usually obtains better precise color reconstruction and vivid color saturation. Meanwhile, our method faithfully reconstructs fine high-frequency textures. For instance, in Fig. \ref{scie}, the newest method, CoTF, exhibits distortion and color cast, but our LALNet still performs well. In Fig. \ref{hdrplus}, our method exhibits excellent color fidelity and restores proper global brightness and local contrast, consistent colors, and sharp details. These results prove that our method produces more pleasing visual effects. 
More results and visual comparisons are presented in our Appendix and \href{https://xxxxxx2025.github.io/LALNet/}{LALNet}.
\begin{table}[ht]
    \centering
    \vspace{-0.2cm}
    \caption{Quantitative results of LLE methods on the LOLv1 dataset. "*" denotes that the results are from reference papers.}
    \label{tab:LLE}
    \scalebox{0.76}{
    \begin{tabular}{c|c|cc}
        \toprule[1pt]
        \multirow{2}{*}{Method} & \multirow{2}{*}{GFLOPs} &  \multicolumn{2}{c}{Low-Light Enhancement}  \\
    \cline{3-4}
     &  & PSNR$\uparrow$ & SSIM$\uparrow$   \\ 
    \midrule
    DeepUPE  ~\citep{wang2019underexposed}     & 21.10 &  14.38 &  0.446 \\
    DeepLPF ~\citep{moran2020deeplpf}      & 5.86   &  15.28  &  0.473             \\
    UFormer ~\citep{wang2022uformer}      & 12.00  &  16.36 &  0.771            \\
    RentinexNet ~\citep{wei2018deep}  & 587.47 &  17.19 &  0.589               \\
    EnGAN ~\citep{jiang2021deep}        & 61.01  &  17.48 &  0.650             \\
    Sparse  ~\citep{yang2021sparse}    & 53.26  &  17.20 &  0.640             \\
    FIDE ~\citep{xu2020learning}        & 28.51  &  18.27 &  0.665             \\
    KinD ~\citep{zhang2019kindling}         & 34.99  &  20.35 &  0.813             \\
    MIRNet ~\citep{zamir2020learning}  & 785   &  {\color{blue}\underline{24.14}} &  0.842\\
    LANet  ~\citep{Yang_Cheng_2023}      & /  &  21.71 &  0.810              \\
    Restormer ~\citep{zamir2022restormer}    & 144.25   & 22.43  & 0.823          \\     
    CoTF  ~\citep{li2024cotf}  & 1.81   & 20.06  & 0.755 \\
    Retinexformer ~\citep{cai2023retinexformer}*& 15.57  & 23.93  &  0.831          \\
    Diff-Retinex ~\citep{yi2023diff} & 396.32  & 21.98  &  {\color{blue}\underline{0.852}} \\
    DiffIR ~\citep{xia2023diffir} & 51.63  & 23.15  &  0.828\\
    RetinexMamba ~\citep{bai2024retinexmamba} & 42.82  &  24.03 &  0.827            \\
    \midrule 
    LALNet-Tiny  & \textbf{1.75}   & 24.07   & 0.845 \\
    LALNet & \textbf{6.70}  & \textbf{{\color{red}25.26}} & \textbf{{\color{red}0.855}} \\  
    \bottomrule[1pt]
    \end{tabular}}
    \vspace{-0.5cm}
\end{table}

\subsection{Ablation studies}
We conduct comprehensive breakdown ablations to evaluate the effects of our proposed framework. 

\textbf{Effectiveness of specific modules.} 
To validate the effectiveness of the MCM, DDCM, LGA, and LSSM modules, we set up different variants to validate the effectiveness of the proposed framework. The results are listed in Tab. \ref{tab:keymodule}. 
Variants \#1 serve as the baseline model and represent the removal of all modules and replacement with residual blocks. 
For Variants \#2 apply a convolution block to replace the MCM with a performance reduction of 0.65 dB PSNR. 
In Variants \#3, we use group convolution replacing DDCM to extract channel-separated features, and the PSNR is reduced by 1.16 dB. 
For Variants \#4, we remove the LGA module and directly sum channel-mixed and channel-separated features for light guidance. The results confirm the effectiveness of the color-separated feature to guide the light adaptation, with a PSNR increase of 1.11 dB. 
For Variants \#5, we replace the LSSM module with residual blocks and the performance drops by 1.82 dB.
The results show that our proposed DDCM, LSSM, LGA, and MCM are effective compared to conventional feature extraction. These results consistently demonstrate the effectiveness of our method.
\begin{table}[t]
    \centering
    \caption{Ablation studies of key components on SCIE dataset.}
    \label{tab:keymodule}
    \scalebox{0.82}{
    \begin{tabular}{c|ccccccc}
        \toprule[1pt]
        {Variants} & {MCM} & {DDCM} & {LGA} & {LSSM} & PSNR$\uparrow$ & SSIM$\uparrow$\\
        \midrule
        \#1     & \XSolidBrush & \XSolidBrush & \XSolidBrush & \XSolidBrush & 20.29	& 0.6834 \\
        \#2     & \XSolidBrush & \Checkmark & \Checkmark  & \Checkmark & 23.95	& 0.7137 \\
        \#3     & \Checkmark & \XSolidBrush & \Checkmark & \Checkmark & 23.47	& 0.7144 \\
        \#4     & \Checkmark & \Checkmark & \XSolidBrush & \Checkmark & 23.53	& 0.7229 \\ 
        \#5     & \Checkmark & \Checkmark & \Checkmark  & \XSolidBrush & 22.81	& 0.7091  \\
        \#6     & \Checkmark & \Checkmark & \Checkmark  & \Checkmark & 24.63  &  0.7270 \\
        \bottomrule[1pt]
    \end{tabular}
    }
    \vspace{-0.5cm}
\end{table}

\begin{table}[ht]
    \centering
    \caption{Ablation study on the pyramid levels number. The "N.A." result is not available due to insufficient GPU memory.}
    \label{tab:level}
    \scalebox{0.92}{
    \begin{tabular}{c|cccc}
    \toprule[1pt]
    \multirow{2}{*}{Metrics} & \multicolumn{4}{c}{Number of Levels} \\
    \cline{2-5}
    & n=1  & n=2  & n=3  & n=4 \\ 
    \hline
    PSNR           & N.A. & 23.45 & \textbf{24.63} & 23.07  \\
    SSIM           & N.A. & 0.7102 & \textbf{0.7270} & 0.7094   \\
    TMQI           & N.A. & 0.8735 & \textbf{0.8667} & 0.8783  \\
    LPIPS          & N.A. & 0.1240 & \textbf{0.1270} & 0.116   \\
    $\bigtriangleup$E  & N.A. & 8.18  & \textbf{7.53}  & 8.22    \\
    \#Params         & \textbf{2.20M} & 2.29M  & 2.45M  & 2.81M  \\
    FLOPs        & 36.16G & 12.20G  & 6.70G  & \textbf{5.51G}  \\
        \bottomrule[1pt]
    \end{tabular}
    }
    \vspace{-0.5cm}
\end{table}

\textbf{Selection of the number of levels.} 
We validate the influence of the number of pyramid levels $l$. As shown in Tab. \ref{tab:level}, the model achieves the best performance on all tested resolutions when $l=3$. When a larger number of levels ($l\geq4$) results in a significant decline in performance. This is because when $l$ is larger and the number of downsamples is more, the model fails to reconstruct the high frequencies efficiently, resulting in performance degradation. When $l=1$, the low-frequency image resolution equals the input image resolution, leading to a burst of computational memory. Comparing $l=2$ and $l=3$ demonstrates that despite the small input image resolution of the low-frequency pathway, high-frequency details can still be recovered efficiently in our framework.

\section{Conclusion}
\label{conc}
This paper proposes a unified framework for learning adaptive lighting via light property guidance. In particular, we propose DDCM for extracting color-separated features and capturing the light difference across channels. The LGA utilizes color-separated features to guide color-mixed features for adaptive lighting, achieving color consistency and color balance. Extensive experiments demonstrate that our method significantly outperforms state-of-the-art methods, improving PSNR by 0.86 dB in the SCIE dataset, 2.15 dB in the HDR+ dataset, 1.12 dB in the LOL dataset, and 3.86 dB in the HDRI Haven dataset respectively compared with the second-best method. 

\section*{Acknowledgement}
This work is supported by the National Natural Science Foundation of China under Grants 62231018 and 62072331.

\section*{Impact Statement}
\label{boardimp}
The proposed LALNet framework represents a significant advancement in the field of image processing, offering a versatile and efficient solution for various light-related tasks. By adapting the training data, LALNet makes it easier for even non-experts to process high-quality images, contributing positively to areas where image quality is crucial. In addition, this research can further inspire developments in areas such as autonomous driving, where complex problems are solved in novel ways.

\bibliography{ref}

\begin{thebibliography}{72}
\providecommand{\natexlab}[1]{#1}
\providecommand{\url}[1]{\texttt{#1}}
\expandafter\ifx\csname urlstyle\endcsname\relax
  \providecommand{\doi}[1]{doi: #1}\else
  \providecommand{\doi}{doi: \begingroup \urlstyle{rm}\Url}\fi

\bibitem[Afifi et~al.(2021)Afifi, Derpanis, Ommer, and Brown]{afifi2021learning}
Afifi, M., Derpanis, K.~G., Ommer, B., and Brown, M.~S.
\newblock Learning multi-scale photo exposure correction.
\newblock In \emph{Proceedings of the IEEE/CVF Conference on Computer Vision and Pattern Recognition}, pp.\  9157--9167, 2021.

\bibitem[Bai et~al.(2024)Bai, Yin, and He]{bai2024retinexmamba}
Bai, J., Yin, Y., and He, Q.
\newblock Retinexmamba: Retinex-based mamba for low-light image enhancement.
\newblock \emph{arXiv preprint arXiv:2405.03349}, 2024.

\bibitem[Cai et~al.(2018)Cai, Gu, and Zhang]{cai2018learning}
Cai, J., Gu, S., and Zhang, L.
\newblock Learning a deep single image contrast enhancer from multi-exposure images.
\newblock \emph{IEEE Transactions on Image Processing}, 27\penalty0 (4):\penalty0 2049--2062, 2018.

\bibitem[Cai et~al.(2023)Cai, Bian, Lin, Wang, Timofte, and Zhang]{cai2023retinexformer}
Cai, Y., Bian, H., Lin, J., Wang, H., Timofte, R., and Zhang, Y.
\newblock Retinexformer: One-stage retinex-based transformer for low-light image enhancement.
\newblock In \emph{Proceedings of the IEEE/CVF International Conference on Computer Vision}, pp.\  12504--12513, 2023.

\bibitem[Cao et~al.(2023)Cao, Yue, Liu, and Yang]{cao2023unsupervised}
Cao, C., Yue, H., Liu, X., and Yang, J.
\newblock Unsupervised hdr image and video tone mapping via contrastive learning.
\newblock \emph{IEEE Transactions on Circuits and Systems for Video Technology}, 2023.

\bibitem[Cao et~al.(2020)Cao, Lai, Yanushkevich, and Smith]{Cao_2020}
Cao, X., Lai, K., Yanushkevich, S., and Smith, M.~R.
\newblock Adversarial and adaptive tone mapping operator for high dynamic range images.
\newblock In \emph{2020 IEEE Symposium Series on Computational Intelligence (SSCI)}, Dec 2020.

\bibitem[Chen et~al.(2018)Chen, Wang, Kao, and Chuang]{chen2018deep}
Chen, Y.-S., Wang, Y.-C., Kao, M.-H., and Chuang, Y.-Y.
\newblock Deep photo enhancer: Unpaired learning for image enhancement from photographs with gans.
\newblock In \emph{Proceedings of the IEEE conference on computer vision and pattern recognition}, pp.\  6306--6314, 2018.

\bibitem[Fairchild(2023)]{Fairchild_2023}
Fairchild, M.~D.
\newblock The hdr photographic survey.
\newblock \emph{Color and Imaging Conference}, pp.\  233–238, May 2023.
\newblock \doi{10.2352/cic.2007.15.1.art00044}.
\newblock URL \url{http://dx.doi.org/10.2352/cic.2007.15.1.art00044}.

\bibitem[Finder et~al.(2024)Finder, Amoyal, Treister, and Freifeld]{finder2024wavelet}
Finder, S.~E., Amoyal, R., Treister, E., and Freifeld, O.
\newblock Wavelet convolutions for large receptive fields.
\newblock \emph{arXiv preprint arXiv:2407.05848}, 2024.

\bibitem[Fu et~al.(2016)Fu, Zeng, Huang, Zhang, and Ding]{fu2016weighted}
Fu, X., Zeng, D., Huang, Y., Zhang, X.-P., and Ding, X.
\newblock A weighted variational model for simultaneous reflectance and illumination estimation.
\newblock In \emph{Proceedings of the IEEE conference on computer vision and pattern recognition}, pp.\  2782--2790, 2016.

\bibitem[Gao \& Wu(2021)Gao and Wu]{gao2021real}
Gao, Q. and Wu, X.
\newblock Real-time deep image retouching based on learnt semantics dependent global transforms.
\newblock \emph{IEEE Transactions on Image Processing}, 30:\penalty0 7378--7390, 2021.

\bibitem[Gharbi et~al.(2017)Gharbi, Chen, Barron, Hasinoff, and Durand]{Gharbi_hdrnet_2017}
Gharbi, M., Chen, J., Barron, J.~T., Hasinoff, S.~W., and Durand, F.
\newblock Deep bilateral learning for real-time image enhancement.
\newblock \emph{ACM Transactions on Graphics}, pp.\  1–12, Aug 2017.

\bibitem[Guo et~al.(2020)Guo, Li, Guo, Loy, Hou, Kwong, and Cong]{guo2020zero}
Guo, C., Li, C., Guo, J., Loy, C.~C., Hou, J., Kwong, S., and Cong, R.
\newblock Zero-reference deep curve estimation for low-light image enhancement.
\newblock In \emph{Proceedings of the IEEE/CVF conference on computer vision and pattern recognition}, pp.\  1780--1789, 2020.

\bibitem[Guo et~al.(2024)Guo, Li, Dai, Ouyang, Ren, and Xia]{guo2024mambair}
Guo, H., Li, J., Dai, T., Ouyang, Z., Ren, X., and Xia, S.-T.
\newblock Mambair: A simple baseline for image restoration with state-space model.
\newblock \emph{arXiv preprint arXiv:2402.15648}, 2024.

\bibitem[Hasinoff et~al.(2016)Hasinoff, Sharlet, Geiss, Adams, Barron, Kainz, Chen, and Levoy]{Hasinoff_hdrplus_2016}
Hasinoff, S.~W., Sharlet, D., Geiss, R., Adams, A., Barron, J.~T., Kainz, F., Chen, J., and Levoy, M.
\newblock Burst photography for high dynamic range and low-light imaging on mobile cameras.
\newblock \emph{ACM Transactions on Graphics}, pp.\  1–12, Nov 2016.

\bibitem[He et~al.(2020)He, Liu, Qiao, and Dong]{He_CSRnet_2020}
He, J., Liu, Y., Qiao, Y., and Dong, C.
\newblock Conditional sequential modulation for efficient global image retouching.
\newblock In \emph{Computer Vision--ECCV 2020: 16th European Conference, Glasgow, UK, August 23--28, 2020, Proceedings, Part XIII 16}, pp.\  679--695. Springer, 2020.

\bibitem[Hou et~al.(2017)Hou, Duan, and Qiu]{hou2017deep}
Hou, X., Duan, J., and Qiu, G.
\newblock Deep feature consistent deep image transformations: Downscaling, decolorization and hdr tone mapping.
\newblock \emph{arXiv preprint arXiv:1707.09482}, 2017.

\bibitem[Hu et~al.(2022)Hu, Chen, and Allebach]{Hu_2022}
Hu, L., Chen, H., and Allebach, J.~P.
\newblock Joint multi-scale tone mapping and denoising for hdr image enhancement.
\newblock In \emph{2022 IEEE/CVF Winter Conference on Applications of Computer Vision Workshops (WACVW)}, Jan 2022.

\bibitem[Huang et~al.(2022{\natexlab{a}})Huang, Liu, Fu, Zhou, Wang, Zhao, and Xiong]{huang2022exposure}
Huang, J., Liu, Y., Fu, X., Zhou, M., Wang, Y., Zhao, F., and Xiong, Z.
\newblock Exposure normalization and compensation for multiple-exposure correction.
\newblock In \emph{Proceedings of the IEEE/CVF Conference on Computer Vision and Pattern Recognition}, pp.\  6043--6052, 2022{\natexlab{a}}.

\bibitem[Huang et~al.(2022{\natexlab{b}})Huang, Liu, Zhao, Yan, Zhang, Huang, Zhou, and Xiong]{huang2022deep}
Huang, J., Liu, Y., Zhao, F., Yan, K., Zhang, J., Huang, Y., Zhou, M., and Xiong, Z.
\newblock Deep fourier-based exposure correction network with spatial-frequency interaction.
\newblock In \emph{European Conference on Computer Vision}, pp.\  163--180. Springer, 2022{\natexlab{b}}.

\bibitem[Huang et~al.(2023)Huang, Zhao, Zhou, Xiao, Zheng, Zheng, and Xiong]{huang2023learning}
Huang, J., Zhao, F., Zhou, M., Xiao, J., Zheng, N., Zheng, K., and Xiong, Z.
\newblock Learning sample relationship for exposure correction.
\newblock In \emph{Proceedings of the IEEE/CVF conference on computer vision and pattern recognition}, pp.\  9904--9913, 2023.

\bibitem[Jiang et~al.(2021)Jiang, Gong, Liu, Cheng, Fang, Shen, Yang, Zhou, and Enlightengan]{jiang2021deep}
Jiang, Y., Gong, X., Liu, D., Cheng, Y., Fang, C., Shen, X., Yang, J., Zhou, P., and Enlightengan, Z.~W.
\newblock Deep light enhancement without paired supervision., 2021, 30.
\newblock \emph{DOI: https://doi. org/10.1109/TIP}, pp.\  2340--2349, 2021.

\bibitem[Kim et~al.(2020)Kim, Koh, and Kim]{kim2020global}
Kim, H.-U., Koh, Y.~J., and Kim, C.-S.
\newblock Global and local enhancement networks for paired and unpaired image enhancement.
\newblock In \emph{Computer Vision--ECCV 2020: 16th European Conference, Glasgow, UK, August 23--28, 2020, Proceedings, Part XXV 16}, pp.\  339--354. Springer, 2020.

\bibitem[Li et~al.(2020)Li, Guo, Ai, Zhou, and Loy]{li2020flexible}
Li, C., Guo, C., Ai, Q., Zhou, S., and Loy, C.~C.
\newblock Flexible piecewise curves estimation for photo enhancement.
\newblock \emph{arXiv preprint arXiv:2010.13412}, 2020.

\bibitem[Li et~al.(2024{\natexlab{a}})Li, Zhang, Cao, Zhang, Shao, Wang, and Sang]{li2024cotf}
Li, Z., Zhang, F., Cao, M., Zhang, J., Shao, Y., Wang, Y., and Sang, N.
\newblock Real-time exposure correction via collaborative transformations and adaptive sampling.
\newblock In \emph{Proceedings of the IEEE/CVF Conference on Computer Vision and Pattern Recognition}, pp.\  2984--2994, 2024{\natexlab{a}}.

\bibitem[Li et~al.(2024{\natexlab{b}})Li, Zhang, Cao, Zhang, Shao, Wang, and Sang]{li2024real}
Li, Z., Zhang, F., Cao, M., Zhang, J., Shao, Y., Wang, Y., and Sang, N.
\newblock Real-time exposure correction via collaborative transformations and adaptive sampling.
\newblock In \emph{Proceedings of the IEEE/CVF Conference on Computer Vision and Pattern Recognition}, pp.\  2984--2994, 2024{\natexlab{b}}.

\bibitem[Liang et~al.(2021{\natexlab{a}})Liang, Zeng, Cui, Xie, and Zhang]{liang2021ppr10k}
Liang, J., Zeng, H., Cui, M., Xie, X., and Zhang, L.
\newblock Ppr10k: A large-scale portrait photo retouching dataset with human-region mask and group-level consistency.
\newblock In \emph{Proceedings of the IEEE/CVF Conference on Computer Vision and Pattern Recognition}, pp.\  653--661, 2021{\natexlab{a}}.

\bibitem[Liang et~al.(2021{\natexlab{b}})Liang, Zeng, and Zhang]{Liang_lptn_2021}
Liang, J., Zeng, H., and Zhang, L.
\newblock High-resolution photorealistic image translation in real-time: A laplacian pyramid translation network.
\newblock In \emph{2021 IEEE/CVF Conference on Computer Vision and Pattern Recognition (CVPR)}, Jun 2021{\natexlab{b}}.

\bibitem[Liang et~al.(2018)Liang, Xu, Zhang, Cao, and Zhang]{Liang2018}
Liang, Z., Xu, J., Zhang, D., Cao, Z., and Zhang, L.
\newblock A hybrid l1-l0 layer decomposition model for tone mapping.
\newblock In \emph{2018 IEEE/CVF Conference on Computer Vision and Pattern Recognition}, Jun 2018.

\bibitem[Liang et~al.(2023)Liang, Li, Zhou, Feng, and Loy]{liang2023iterative}
Liang, Z., Li, C., Zhou, S., Feng, R., and Loy, C.~C.
\newblock Iterative prompt learning for unsupervised backlit image enhancement.
\newblock In \emph{Proceedings of the IEEE/CVF International Conference on Computer Vision}, pp.\  8094--8103, 2023.

\bibitem[Liu et~al.(2021{\natexlab{a}})Liu, Ma, Zhang, Fan, and Luo]{liu2021retinex}
Liu, R., Ma, L., Zhang, J., Fan, X., and Luo, Z.
\newblock Retinex-inspired unrolling with cooperative prior architecture search for low-light image enhancement.
\newblock In \emph{Proceedings of the IEEE/CVF conference on computer vision and pattern recognition}, pp.\  10561--10570, 2021{\natexlab{a}}.

\bibitem[Liu et~al.(2021{\natexlab{b}})Liu, Lin, Cao, Hu, Wei, Zhang, Lin, and Guo]{liu2021swin}
Liu, Z., Lin, Y., Cao, Y., Hu, H., Wei, Y., Zhang, Z., Lin, S., and Guo, B.
\newblock Swin transformer: Hierarchical vision transformer using shifted windows.
\newblock In \emph{Proceedings of the IEEE/CVF international conference on computer vision}, pp.\  10012--10022, 2021{\natexlab{b}}.

\bibitem[Moran et~al.(2020)Moran, Marza, McDonagh, Parisot, and Slabaugh]{moran2020deeplpf}
Moran, S., Marza, P., McDonagh, S., Parisot, S., and Slabaugh, G.
\newblock Deeplpf: Deep local parametric filters for image enhancement.
\newblock In \emph{Proceedings of the IEEE/CVF conference on computer vision and pattern recognition}, pp.\  12826--12835, 2020.

\bibitem[Ni et~al.(2020)Ni, Yang, Wang, Ma, and Kwong]{ni2020towards}
Ni, Z., Yang, W., Wang, S., Ma, L., and Kwong, S.
\newblock Towards unsupervised deep image enhancement with generative adversarial network.
\newblock \emph{IEEE Transactions on Image Processing}, 29:\penalty0 9140--9151, 2020.

\bibitem[Nsampi et~al.(2021)Nsampi, Hu, and Wang]{nsampi2021learning}
Nsampi, N.~E., Hu, Z., and Wang, Q.
\newblock Learning exposure correction via consistency modeling.
\newblock In \emph{BMVC}, pp.\ ~12, 2021.

\bibitem[Panetta et~al.(2021)Panetta, Kezebou, Oludare, Agaian, and Xia]{Panetta_2021}
Panetta, K., Kezebou, L., Oludare, V., Agaian, S., and Xia, Z.
\newblock Tmo-net: A parameter-free tone mapping operator using generative adversarial network, and performance benchmarking on large scale hdr dataset.
\newblock \emph{IEEE Access}, pp.\  39500–39517, Jan 2021.

\bibitem[Paris et~al.(2011)Paris, Hasinoff, and Kautz]{Paris2011}
Paris, S., Hasinoff, S.~W., and Kautz, J.
\newblock Local laplacian filters.
\newblock In \emph{ACM SIGGRAPH 2011 papers}, Jul 2011.

\bibitem[Pizer et~al.(1987)Pizer, Amburn, Austin, Cromartie, Geselowitz, Greer, ter Haar~Romeny, Zimmerman, and Zuiderveld]{pizer1987adaptive}
Pizer, S.~M., Amburn, E.~P., Austin, J.~D., Cromartie, R., Geselowitz, A., Greer, T., ter Haar~Romeny, B., Zimmerman, J.~B., and Zuiderveld, K.
\newblock Adaptive histogram equalization and its variations.
\newblock \emph{Computer vision, graphics, and image processing}, 39\penalty0 (3):\penalty0 355--368, 1987.

\bibitem[Rana et~al.(2020)Rana, Singh, Valenzise, Dufaux, Komodakis, and Smolic]{Rana_DEEPTMO_2020}
Rana, A., Singh, P., Valenzise, G., Dufaux, F., Komodakis, N., and Smolic, A.
\newblock Deep tone mapping operator for high dynamic range images.
\newblock \emph{IEEE Transactions on Image Processing}, pp.\  1285–1298, Jan 2020.

\bibitem[Reza(2004)]{reza2004realization}
Reza, A.~M.
\newblock Realization of the contrast limited adaptive histogram equalization (clahe) for real-time image enhancement.
\newblock \emph{Journal of VLSI signal processing systems for signal, image and video technology}, 38:\penalty0 35--44, 2004.

\bibitem[Rieke \& Rudd(2009)Rieke and Rudd]{Rieke_Rudd_2009}
Rieke, F. and Rudd, M.~E.
\newblock The challenges natural images pose for visual adaptation.
\newblock \emph{Neuron}, 64\penalty0 (5):\penalty0 605–616, Dec 2009.
\newblock \doi{10.1016/j.neuron.2009.11.028}.
\newblock URL \url{http://dx.doi.org/10.1016/j.neuron.2009.11.028}.

\bibitem[Shensa et~al.(1992)]{shensa1992discrete}
Shensa, M.~J. et~al.
\newblock The discrete wavelet transform: wedding the a trous and mallat algorithms.
\newblock \emph{IEEE Transactions on signal processing}, 40\penalty0 (10):\penalty0 2464--2482, 1992.

\bibitem[Su et~al.(2021)Su, Wang, Lin, Liu, Chen, Chang, and Pei]{su2021explorable}
Su, C.-C., Wang, R., Lin, H.-J., Liu, Y.-L., Chen, C.-P., Chang, Y.-L., and Pei, S.-C.
\newblock Explorable tone mapping operators.
\newblock In \emph{2020 25th International Conference on Pattern Recognition (ICPR)}, pp.\  10320--10326. IEEE, 2021.

\bibitem[Su et~al.(2024)Su, Wang, Liu, Han, Fu, and Liao]{su2024styleretoucher}
Su, W., Wang, C., Liu, C., Han, F., Fu, H., and Liao, J.
\newblock Styleretoucher: Generalized portrait image retouching with gan priors.
\newblock \emph{IEEE Transactions on Visualization and Computer Graphics}, 2024.

\bibitem[Wang et~al.(2023)Wang, Zhang, Liu, Wu, and Zuo]{wang2023learning}
Wang, H., Zhang, J., Liu, M., Wu, X., and Zuo, W.
\newblock Learning diverse tone styles for image retouching.
\newblock \emph{IEEE Transactions on Image Processing}, 2023.

\bibitem[Wang et~al.(2019{\natexlab{a}})Wang, Zhang, Fu, Shen, Zheng, and Jia]{Wang_UPE_2019}
Wang, R., Zhang, Q., Fu, C.-W., Shen, X., Zheng, W.-S., and Jia, J.
\newblock Underexposed photo enhancement using deep illumination estimation.
\newblock In \emph{2019 IEEE/CVF Conference on Computer Vision and Pattern Recognition (CVPR)}, Jun 2019{\natexlab{a}}.

\bibitem[Wang et~al.(2019{\natexlab{b}})Wang, Zhang, Fu, Shen, Zheng, and Jia]{wang2019underexposed}
Wang, R., Zhang, Q., Fu, C.-W., Shen, X., Zheng, W.-S., and Jia, J.
\newblock Underexposed photo enhancement using deep illumination estimation.
\newblock In \emph{Proceedings of the IEEE/CVF conference on computer vision and pattern recognition}, pp.\  6849--6857, 2019{\natexlab{b}}.

\bibitem[Wang et~al.(2021)Wang, Li, Peng, Ma, Wang, Song, and Yan]{wang2021real}
Wang, T., Li, Y., Peng, J., Ma, Y., Wang, X., Song, F., and Yan, Y.
\newblock Real-time image enhancer via learnable spatial-aware 3d lookup tables.
\newblock In \emph{Proceedings of the IEEE/CVF International Conference on Computer Vision}, pp.\  2471--2480, 2021.

\bibitem[Wang et~al.(2022)Wang, Cun, Bao, Zhou, Liu, and Li]{wang2022uformer}
Wang, Z., Cun, X., Bao, J., Zhou, W., Liu, J., and Li, H.
\newblock Uformer: A general u-shaped transformer for image restoration.
\newblock In \emph{Proceedings of the IEEE/CVF conference on computer vision and pattern recognition}, pp.\  17683--17693, 2022.

\bibitem[Wei et~al.(2018)Wei, Wang, Yang, and Liu]{wei2018deep}
Wei, C., Wang, W., Yang, W., and Liu, J.
\newblock Deep retinex decomposition for low-light enhancement.
\newblock \emph{arXiv preprint arXiv:1808.04560}, 2018.

\bibitem[Wu et~al.(2022)Wu, Weng, Zhang, Wang, Yang, and Jiang]{wu2022uretinex}
Wu, W., Weng, J., Zhang, P., Wang, X., Yang, W., and Jiang, J.
\newblock Uretinex-net: Retinex-based deep unfolding network for low-light image enhancement.
\newblock In \emph{Proceedings of the IEEE/CVF conference on computer vision and pattern recognition}, pp.\  5901--5910, 2022.

\bibitem[Xia et~al.(2023)Xia, Zhang, Wang, Wang, Wu, Tian, Yang, and Van~Gool]{xia2023diffir}
Xia, B., Zhang, Y., Wang, S., Wang, Y., Wu, X., Tian, Y., Yang, W., and Van~Gool, L.
\newblock Diffir: Efficient diffusion model for image restoration.
\newblock In \emph{Proceedings of the IEEE/CVF International Conference on Computer Vision}, pp.\  13095--13105, 2023.

\bibitem[Xu et~al.(2020)Xu, Yang, Yin, and Lau]{xu2020learning}
Xu, K., Yang, X., Yin, B., and Lau, R.~W.
\newblock Learning to restore low-light images via decomposition-and-enhancement.
\newblock In \emph{Proceedings of the IEEE/CVF conference on computer vision and pattern recognition}, pp.\  2281--2290, 2020.

\bibitem[Xu et~al.(2022)Xu, Wang, Fu, and Jia]{xu2022snr}
Xu, X., Wang, R., Fu, C.-W., and Jia, J.
\newblock Snr-aware low-light image enhancement.
\newblock In \emph{Proceedings of the IEEE/CVF conference on computer vision and pattern recognition}, pp.\  17714--17724, 2022.

\bibitem[Yang et~al.(2022)Yang, Jin, Xu, Zhang, Chen, and Liu]{Yang_SepLUT_2022}
Yang, C., Jin, M., Xu, Y., Zhang, R., Chen, Y., and Liu, H.
\newblock Seplut: Separable image-adaptive lookup tables for real-time image enhancement.
\newblock In \emph{European Conference on Computer Vision}, pp.\  201--217. Springer, 2022.

\bibitem[Yang et~al.(2023{\natexlab{a}})Yang, Cheng, Zhao, Yan, Zhang, and Li]{Yang_Cheng_2023}
Yang, K.-F., Cheng, C., Zhao, S.-X., Yan, H.-M., Zhang, X.-S., and Li, Y.-J.
\newblock Learning to adapt to light.
\newblock \emph{International Journal of Computer Vision}, 131\penalty0 (4):\penalty0 1022--1041, 2023{\natexlab{a}}.

\bibitem[Yang et~al.(2023{\natexlab{b}})Yang, Yue, Zhang, Liu, Yang, et~al.]{yang2023learning}
Yang, Q., Yue, H., Zhang, L., Liu, Y., Yang, J., et~al.
\newblock Learning to see low-light images via feature domain adaptation.
\newblock \emph{arXiv preprint arXiv:2312.06723}, 2023{\natexlab{b}}.

\bibitem[Yang et~al.(2024)Yang, Li, Jiang, Cheng, Yu, Liu, Yue, and Yang]{yang2024learning}
Yang, Q., Li, Y., Jiang, P.-T., Cheng, Q., Yu, B., Liu, Y., Yue, H., and Yang, J.
\newblock Learning differential pyramid representation for tone mapping.
\newblock \emph{arXiv preprint arXiv:2412.01463}, 2024.

\bibitem[Yang et~al.(2025)Yang, Cheng, Yue, Zhang, Liu, and Yang]{yang2025learning}
Yang, Q., Cheng, Q., Yue, H., Zhang, L., Liu, Y., and Yang, J.
\newblock Learning to see low-light images via feature domain adaptation.
\newblock \emph{IEEE Transactions on Image Processing}, 2025.

\bibitem[Yang et~al.(2020)Yang, Wang, Fang, Wang, and Liu]{yang2020fidelity}
Yang, W., Wang, S., Fang, Y., Wang, Y., and Liu, J.
\newblock From fidelity to perceptual quality: A semi-supervised approach for low-light image enhancement.
\newblock In \emph{Proceedings of the IEEE/CVF conference on computer vision and pattern recognition}, pp.\  3063--3072, 2020.

\bibitem[Yang et~al.(2021)Yang, Wang, Huang, Wang, and Liu]{yang2021sparse}
Yang, W., Wang, W., Huang, H., Wang, S., and Liu, J.
\newblock Sparse gradient regularized deep retinex network for robust low-light image enhancement.
\newblock \emph{IEEE Transactions on Image Processing}, 30:\penalty0 2072--2086, 2021.

\bibitem[Yeganeh \& Wang(2013)Yeganeh and Wang]{Yeganeh_Wang_2013}
Yeganeh, H. and Wang, Z.
\newblock Objective quality assessment of tone-mapped images.
\newblock \emph{IEEE Transactions on Image Processing}, pp.\  657–667, Feb 2013.

\bibitem[Yi et~al.(2023)Yi, Xu, Zhang, Tang, and Ma]{yi2023diff}
Yi, X., Xu, H., Zhang, H., Tang, L., and Ma, J.
\newblock Diff-retinex: Rethinking low-light image enhancement with a generative diffusion model.
\newblock In \emph{Proceedings of the IEEE/CVF International Conference on Computer Vision}, pp.\  12302--12311, 2023.

\bibitem[Zamir et~al.(2020)Zamir, Arora, Khan, Hayat, Khan, Yang, and Shao]{zamir2020learning}
Zamir, S.~W., Arora, A., Khan, S., Hayat, M., Khan, F.~S., Yang, M.-H., and Shao, L.
\newblock Learning enriched features for real image restoration and enhancement.
\newblock In \emph{Computer Vision--ECCV 2020: 16th European Conference, Glasgow, UK, August 23--28, 2020, Proceedings, Part XXV 16}, pp.\  492--511. Springer, 2020.

\bibitem[Zamir et~al.(2022)Zamir, Arora, Khan, Hayat, Khan, and Yang]{zamir2022restormer}
Zamir, S.~W., Arora, A., Khan, S., Hayat, M., Khan, F.~S., and Yang, M.-H.
\newblock Restormer: Efficient transformer for high-resolution image restoration.
\newblock In \emph{Proceedings of the IEEE/CVF conference on computer vision and pattern recognition}, pp.\  5728--5739, 2022.

\bibitem[Zeng et~al.(2020)Zeng, Cai, Li, Cao, and Zhang]{Zeng_lut_2020}
Zeng, H., Cai, J., Li, L., Cao, Z., and Zhang, L.
\newblock Learning image-adaptive 3d lookup tables for high performance photo enhancement in real-time.
\newblock \emph{IEEE Transactions on Pattern Analysis and Machine Intelligence}, pp.\  1–1, Jan 2020.

\bibitem[Zhang et~al.(2022)Zhang, Zeng, Zhang, and Zhang]{Zhang_CLUT}
Zhang, F., Zeng, H., Zhang, T., and Zhang, L.
\newblock Clut-net: Learning adaptively compressed representations of 3dluts for lightweight image enhancement.
\newblock In \emph{Proceedings of the 30th ACM International Conference on Multimedia}, pp.\  6493--6501, 2022.

\bibitem[Zhang et~al.(2024)Zhang, Tian, Li, Xu, Lu, Gao, and Sang]{Zhang_llfLUT_2023}
Zhang, F., Tian, M., Li, Z., Xu, B., Lu, Q., Gao, C., and Sang, N.
\newblock Lookup table meets local laplacian filter: pyramid reconstruction network for tone mapping.
\newblock \emph{Advances in Neural Information Processing Systems}, 36, 2024.

\bibitem[Zhang et~al.(2019{\natexlab{a}})Zhang, Wang, Zhao, and Wang]{zhang2019deep}
Zhang, N., Wang, C., Zhao, Y., and Wang, R.
\newblock Deep tone mapping network in hsv color space.
\newblock In \emph{2019 IEEE Visual Communications and Image Processing (VCIP)}, pp.\  1--4. IEEE, 2019{\natexlab{a}}.

\bibitem[Zhang et~al.(2018)Zhang, Isola, Efros, Shechtman, and Wang]{zhang2018unreasonable}
Zhang, R., Isola, P., Efros, A.~A., Shechtman, E., and Wang, O.
\newblock The unreasonable effectiveness of deep features as a perceptual metric.
\newblock In \emph{Proceedings of the IEEE conference on computer vision and pattern recognition}, pp.\  586--595, 2018.

\bibitem[Zhang et~al.(1996)Zhang, Wandell, et~al.]{zhang1996spatial}
Zhang, X., Wandell, B.~A., et~al.
\newblock A spatial extension of cielab for digital color image reproduction.
\newblock In \emph{SID international symposium digest of technical papers}, volume~27, pp.\  731--734. Citeseer, 1996.

\bibitem[Zhang et~al.(2019{\natexlab{b}})Zhang, Zhang, and Guo]{zhang2019kindling}
Zhang, Y., Zhang, J., and Guo, X.
\newblock Kindling the darkness: A practical low-light image enhancer.
\newblock In \emph{Proceedings of the 27th ACM international conference on multimedia}, pp.\  1632--1640, 2019{\natexlab{b}}.

\end{thebibliography}
\bibliographystyle{icml2025}

\newpage
\clearpage
\appendix
\onecolumn



%
\begin{center}
	\Large\textbf{{Appendix}}\\
\end{center}

In this Appendix, we provide additional results and analysis.

\section{Further Analysis of Motivation}
\label{sec:motivation}
Different wavelengths of light exhibit different response characteristics when an image sensor captures photons for photoelectric conversion. After processing by an image signal processor, these differential responses are sometimes amplified or minimized but are difficult to eliminate. In addition, the differences in the Bayer pattern of different image sensors also result in different channels showing different responses to luminance and noise. Meanwhile, light sources in natural scenes are usually non-uniform, which also leads to the fact that sunlight, shadows, reflections, and other factors can cause RGB channels to respond differently to the same scene.

\begin{figure}[htbp]
    \centering
    \includegraphics[width=0.99\textwidth]{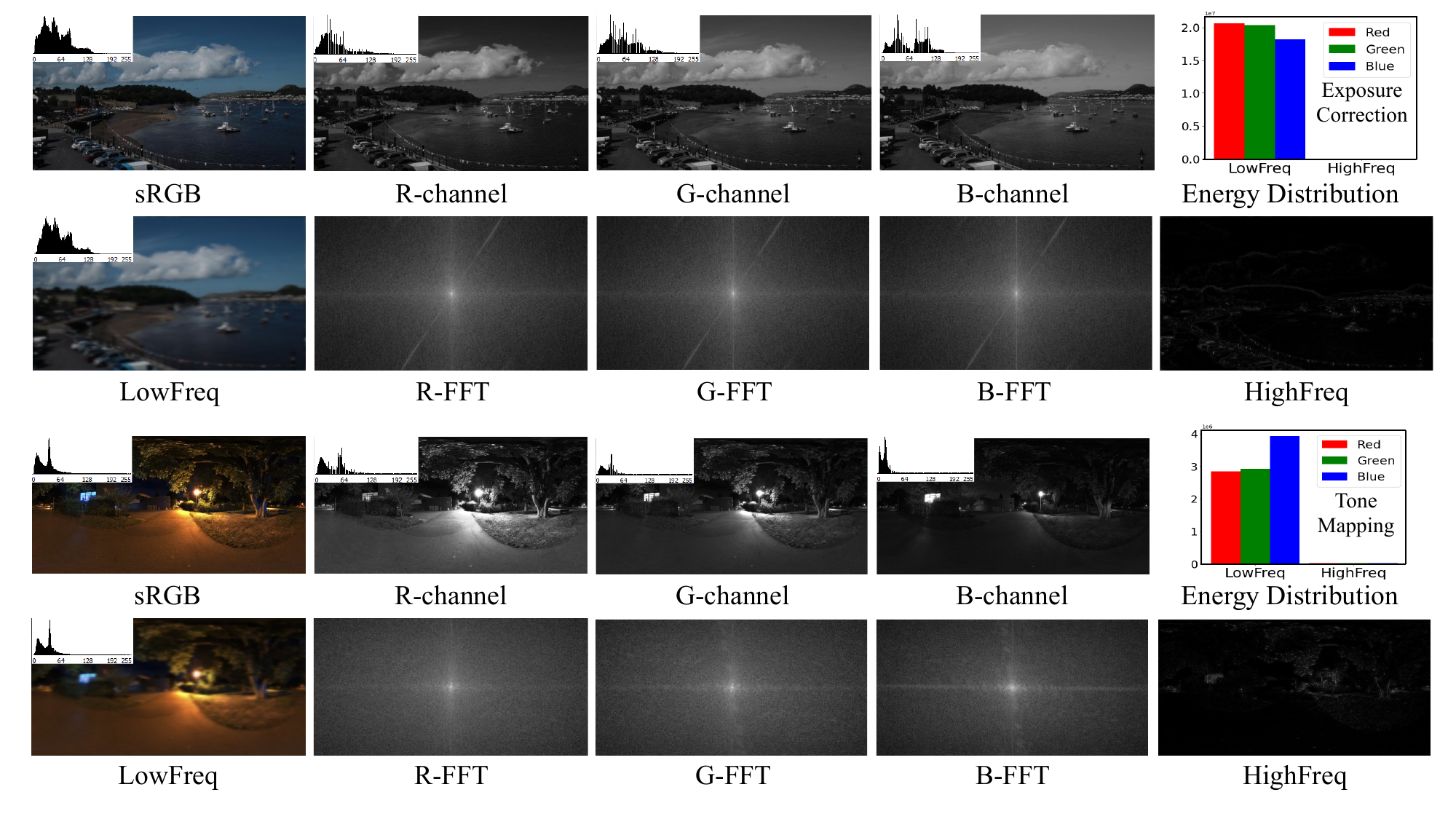}
    \vspace{-0.4cm}
    \caption{Motivation. Visualization of the light-related task images in different color channels and their corresponding DWT spectra energy distribution. R-FFT denotes the Fourier Frequency Domain diagram of the R channel. LowFreq and HighFreq are low-frequency and high-frequency images.}
\label{supmotivation}
\vspace{-0.3cm}
\end{figure}

Recall that in Sec.~\ref{intro}, we discussed two observations that serve as the motivation to design LALNet. We show more motivation cases in Fig. \ref{supmotivation} (the exposure correction and tone mapping tasks). In particular, (a) different color channels have different light properties, and (b) the channel differences reflected in the time and frequency domains are different. To further analyze our first motivation, we visualized the frequency domain images of the different channels using the Fourier Transform and compared them. The results show that, as in the time domain, significant differences are exhibited between the different channels in the frequency domain. Based on the observations in Fig. \ref{motivate} and Fig. \ref{supmotivation}, the common properties of several light-related tasks investigated in this paper are verified, which also contribute to the design of our network.

\vspace{-0.2cm}
\section{Visualization in the Network}
\label{sec:visualization}
\begin{figure}[t]
    \centering
    
    \includegraphics[width=0.6\textwidth]{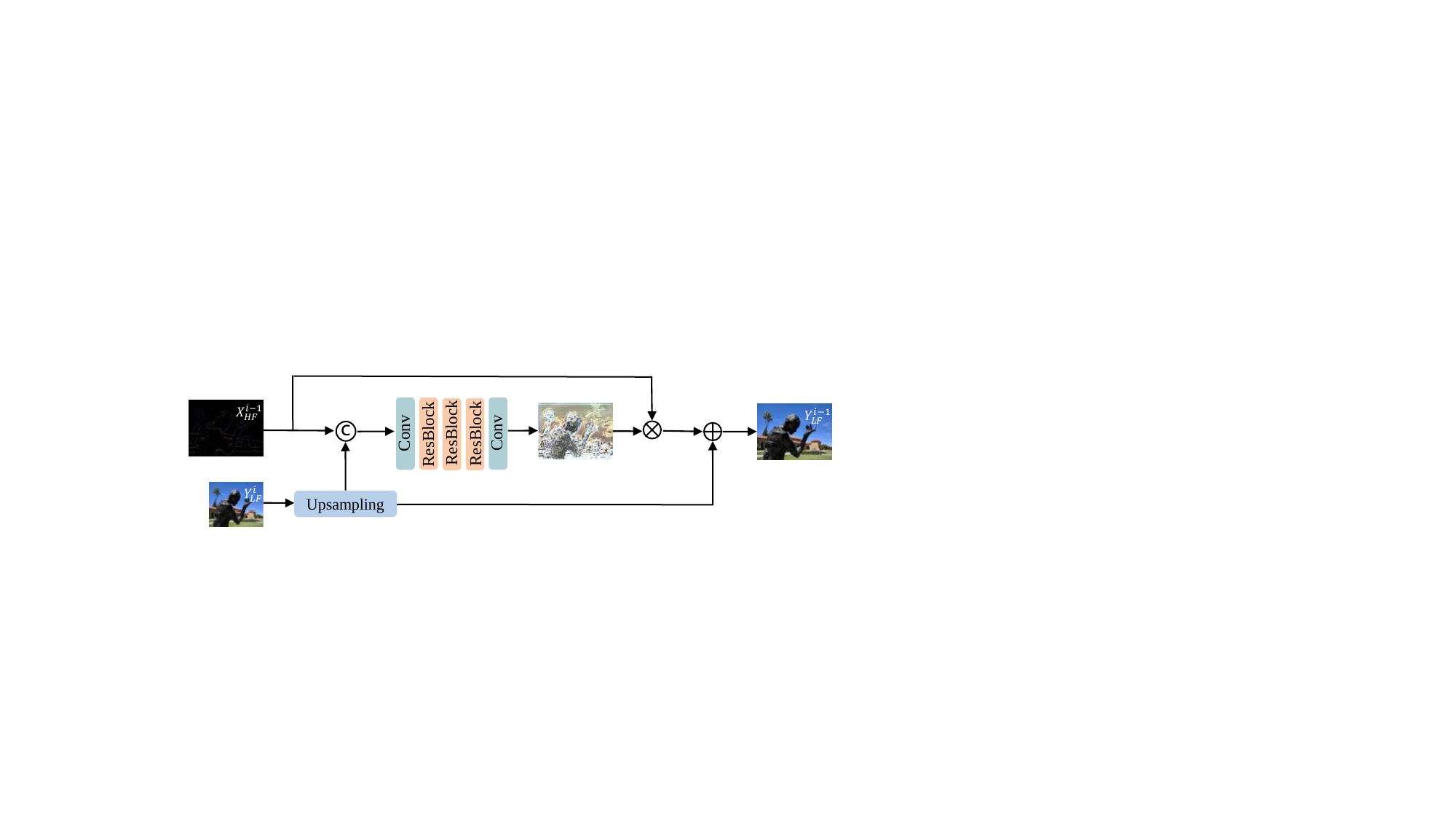}
    \vspace{-0.4cm}
    \caption{The architecture of the Iterative Detail Enhancement module progressively restores resolution and fine details.}
    \vspace{-0.1cm}
\label{ide}
\end{figure}

\begin{figure}[t]
    \centering
    \includegraphics[width=0.55\textwidth]{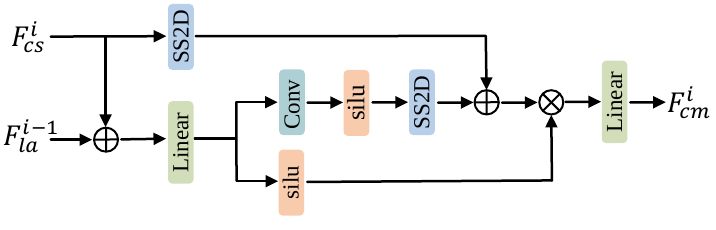}
    \vspace{-0.4cm}
    \caption{The architecture of the Light State Space Module.}
    \vspace{-0.1cm}
\label{lssm}
\end{figure}

We demonstrate the Iterative Detail Enhancement modules (IDE), and Light State Space Module (LSSM) in Fig. \ref{ide} and Fig. \ref{lssm}. 
To reduce the computational resources, we implement light adaptation at low resolution. To compensate for the loss of details, we use an iterative detail enhancement module to recover high-frequency details. 
Specifically, as shown in Fig. \ref{ide}, we first up-sample the low-frequency mapped image $\mathbf{Y}^i_{LF}$ and concatenate it with the HF component $\mathbf{X}^{i-1}_{HF}$, then feed it into a residual network to predict the mask $\mathbf{M}_{i-1}$. This mask allows pixel-by-pixel refinement of the HF component, which is subsequently added to the up-sampling $\mathbf{Y}^i_{LF}$ to generate the reconstructed result of the current layer $\mathbf{Y}^{i-1}_{LF}$. 

\section{More Results}
\label{sec:results}
We further evaluate the effectiveness of our model on the exposure correction~\citep{afifi2021learning}, HDR Survey~\citep{Fairchild_2023}, and UVTM~\citep{cao2023unsupervised} datasets, all of which present more complex lighting conditions.

The MSEC dataset~\citep{afifi2021learning} provides images rendered with relative exposure values (EVs) ranging from -1.5 to +1.5, comprising 17,675 training images, 750 validation images, and 5,905 test images. Table~\ref{tab:MSCE} presents the quantitative results on MSEC. As shown, our method achieves the best overall performance, with a PSNR of 23.93 dB, SSIM of 0.8734, and LPIPS of 0.0791.

\begin{table}[htbp]
\centering
\vspace{-0.4cm}
\caption{Quantitative results of exposure correction methods on the MSCE dataset.}
\label{tab:MSCE}
\small
\scalebox{1}{
\begin{tabular}{c|cc|cc|ccc}
\hline 
\toprule[1pt]
    \multirow{3}{*}{Method} & \multicolumn{7}{c}{Exposure Correction in MSCE}  \\
    \cline{2-8}
    & \multicolumn{2}{c|}{Under} & \multicolumn{2}{c|}{Over} & \multicolumn{3}{c}{Average} \\
     &  PSNR$\uparrow$ & SSIM$\uparrow$  &  PSNR$\uparrow$ & SSIM$\uparrow$ &  PSNR$\uparrow$ & SSIM$\uparrow$ & LPIPS$\downarrow $   \\ 
    \midrule
    He ~\citep{pizer1987adaptive} & 16.52 & 0.6918 & 16.53 & 0.6991 & 16.53 & 0.6959 & 0.2920\\
    CLAHE ~\citep{reza2004realization} & 16.77 & 0.6211 & 14.45 & 0.5842 & 15.38 & 0.5990 & 0.4744\\
    WVM ~\citep{fu2016weighted} & 18.67 & 0.7280 & 12.75 & 0.645 & 15.12 & 0.6780 & 0.2284\\
    RetinexNet ~\citep{wei2018deep}    & 12.13 & 0.6209 & 10.47 & 0.5953 & 11.14 & 0.6048 & 0.3209  \\
    URtinexNet  ~\citep{wu2022uretinex}    & 13.85 & 0.7371 & 9.81 & 0.6733 & 11.42 & 0.6988 & 0.2858  \\
    DRBN  ~\citep{yang2020fidelity}    & 19.74 & 0.8290 & 19.37 & 0.8321 & 19.52 & 0.8309 & 0.2795 \\
    SID ~\citep{chen2018deep}      & 19.37 & 0.8103 & 18.83 & 0.8055 & 19.04 & 0.8074 & 0.1862 \\
    MSEC ~\citep{afifi2021learning}     & 20.52 & 0.8129 & 19.79 & 0.8156 & 20.08 & 0.8145 & 0.1721  \\
    SID-ENC ~\citep{huang2022exposure}      & 22.59 & 0.8423 & 22.36 & 0.8519 & 22.45 & 0.8481 & 0.1827  \\
    DRBN-ENC ~\citep{huang2022exposure}     & 22.72 & 0.8544 & 22.11 & 0.8521 & 22.35 & 0.8530 & 0.1724  \\
    CLIP-LIT ~\citep{liang2023iterative} & 17.79 & 0.7611 & 12.02 & 0.6894 & 14.32 & 0.7181 & 0.2506\\
    FECNet ~\citep{huang2022deep}          & 22.96 & 0.8598 & 23.22 & 0.8748 & 23.12 & 0.8688 & 0.1419  \\
    LCDPNet ~\citep{zhang2019kindling}    & 22.35 & 0.8650 & 22.17 & 0.8476 & 22.30 & 0.8552 & 0.1451   \\
    FECNet+ERL ~\citep{zamir2020learning}   & 23.10 & 0.8639 & 23.18 & 0.8759 & 23.15 & 0.8711 & /  \\
    CoTF  ~\citep{Yang_Cheng_2023}     & 23.36 & 0.8630 & 23.49 & 0.8793 & 23.44 & 0.8728 & 0.1232   \\
    \midrule
    LALNet   &  \textbf{{\color{red}23.78}}	&  \textbf{{\color{red}0.8638}}	& \textbf{{\color{red}24.01}}	& \textbf{{\color{red}0.8787}}	& \textbf{{\color{red}23.90}} & \textbf{{\color{red}0.8713}}	& \textbf{{\color{red}0.0801}} \\
    \bottomrule[1pt]
\end{tabular}}
\end{table}

\begin{table}[htbp]
\centering
\vspace{-0.2cm}
\caption{Validating generalization on third-party datasets, including HDR Survey and UVTM video datasets.}
    \scalebox{0.95}{
    \begin{tabular}{lccccccccccc}
    \toprule[1pt]
        Datasets & Metrics & HDRNet & CSRNet &3D LUT & CLUT & SepLUT  & IVTMNet & CoTF  &Ours \\    
        \midrule
        HDR Survey  & TMQI &0.8641&0.8439 &0.8165 &0.8140 &0.8085 &0.9160 &0.8612  & \textbf{{\color{red}0.9296}}   \\ 
        UVTM & TMQI &0.8281 &0.8973 &0.8787 & 0.8799 &0.8629 &0.8991 &0.9006  & {\color{blue}\underline{0.9584}} \\
        \bottomrule[1pt]
    \end{tabular}}
    \label{tmqi}
\end{table}

\begin{figure*}[t]
    \centering
    \includegraphics[width=0.95\textwidth]{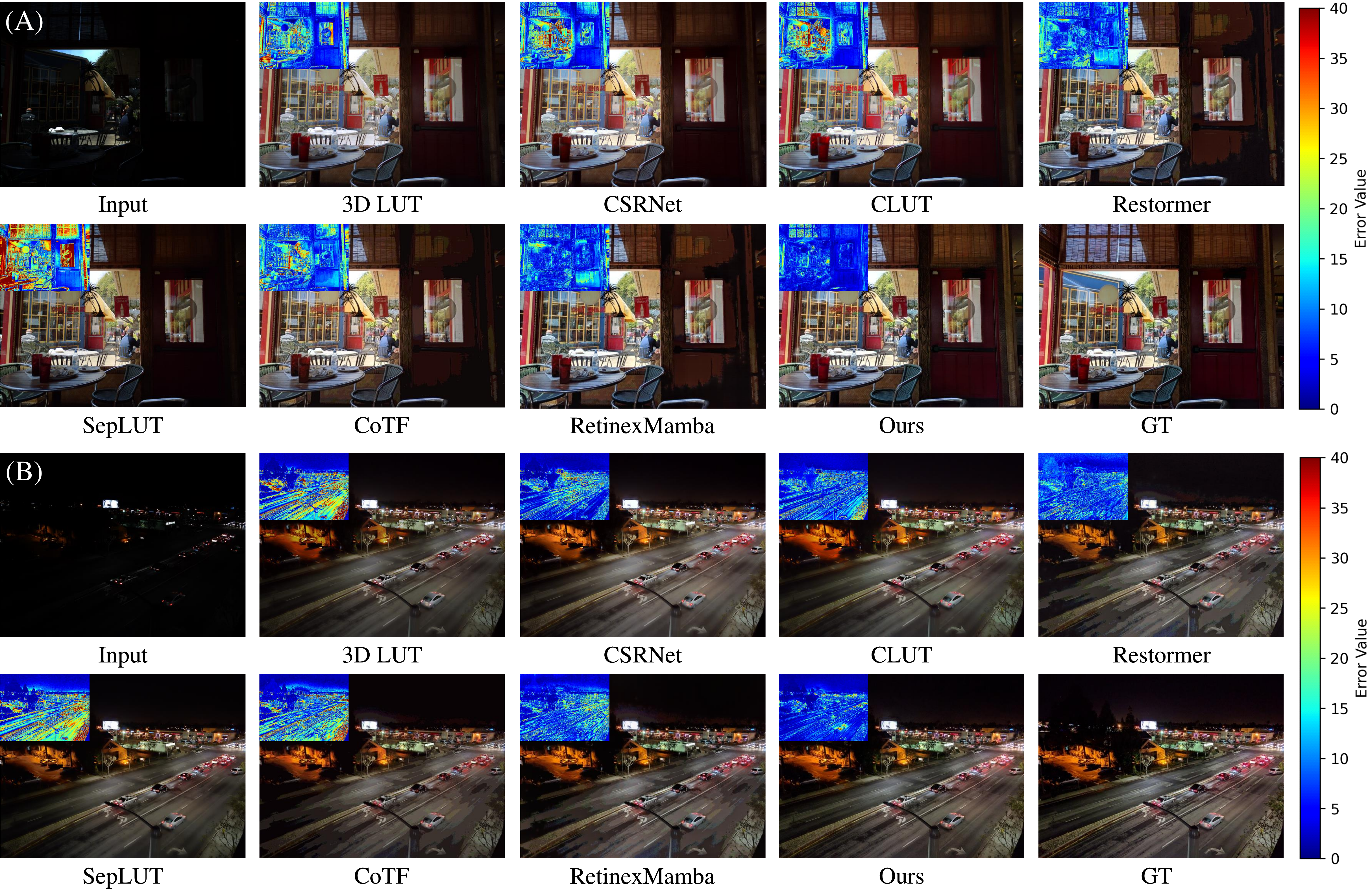}
    \vspace{-0.5cm}
    \caption{Visual comparisons between our LALet and the state-of-the-art methods on the HDR+ dataset.}
    \vspace{-0.2cm}
\label{hdrplus}
\end{figure*}

To further demonstrate the robustness and generalization ability of our model, we evaluate it on third-party non-homologous HDR image (HDR Survey) and video (UVTM) datasets, as summarized in Table~\ref{tmqi}. The HDR Survey dataset comprises 105 HDR images and is widely adopted for benchmarking HDR tone mapping methods~\citep{Cao_2020, Rana_DEEPTMO_2020, Panetta_2021, Liang2018, Paris2011}, though it does not provide ground-truth references. Likewise, the UVTM dataset contains 20 real-world HDR videos, also without ground truth. It is important to note that both the HDR Survey and UVTM datasets are used solely for testing purposes. As shown in Table~\ref{tmqi}, our method significantly outperforms existing approaches on both benchmarks in terms of TMQI. Specifically, our model achieves a TMQI score of 0.9296 on the HDR Survey and 0.9584 on the UVTM, surpassing all competing methods.

\section{Ablation Study}
\label{supsec:ablated}
To validate the effectiveness of the SS2D module, we use Self-Attention and Residual Block to replace the SS2D module in the original published model. We use the Self-Attention module released by Restormer~\citep{zamir2022restormer}, and ResBlock is constructed from two convolutional layers and activation functions. The results, as shown in Table \ref{ss2d}, show that using SS2D as part of the base module effectively captures global features and strikes a balance between performance and efficiency. Notably, the same excellent results are obtained using the Self-Attention module, which is attributed to the design of our overall framework, further demonstrating the effectiveness of our proposed adaptive lighting framework.

Further, we use DDCM to capture color-separated features, and to avoid channel mixing during information propagation, we use group convolution to keep the color channels separated. To verify the effectiveness of the design, we use traditional convolution to replace group convolution. The experimental results are shown in Table \ref{gconv}, where the channel mixing caused by the conventional convolution leads to a performance degradation. This phenomenon shows the necessity of color channel separation and the effectiveness of using color-separated features to guide light adaptation.

\begin{table}[htbp]
\centering
\vspace{-0.3cm}
\caption{Ablation study on the LSSM modules.}
\label{ss2d}
\scalebox{1}{
\begin{tabular}{c|cccccccc} 
\toprule[1pt]
    Variants & Replaced Modules & {\#Params} & {FLOPs} & PSNR$\uparrow$ & SSIM$\uparrow$  & TMQI$\uparrow$  & LPIPS$\downarrow $ & $\bigtriangleup$E$\downarrow $ \\
    \midrule
    \#1           & ResBlock & 2.99M & 7.13G  & 22.81 & 0.7091 & 0.8635 & 0.1291 & 8.480  \\
    \#2           & Self-Attention & 2.25M & 6.48G  & 24.41 & \textbf{0.7253} & 0.8657 & \textbf{0.1257}  & \textbf{7.525} \\
    \midrule 
    \#3           & Ours & 2.45M & 6.70G &\textbf{24.62}	& 0.7227	& \textbf{0.8667}	& 0.1297	& 7.529\\
   \bottomrule[1pt]
\end{tabular}}
\vspace{-0.1cm}
\end{table}

\begin{table}[htbp]
\centering
\vspace{-0.3cm}
\caption{Ablation study on the group Convolution (G-Conv) and traditional Convolution (T-Conv).}
\label{gconv}
\scalebox{1}{
\begin{tabular}{c|cccccccc} 
\toprule[1pt]
    Variants & Replaced Modules & {\#Params} & {FLOPs} & PSNR$\uparrow$ & SSIM$\uparrow$  & TMQI$\uparrow$  & LPIPS$\downarrow $ & $\bigtriangleup$E$\downarrow $ \\
    \midrule
    \#1           & T-Conv & 2.50M & 6.73G & 23.98 & 0.7121 & 0.8521 & 0.1363  & 8.146 \\
    \midrule
    \#2           & G-Conv & 2.45M & 6.70G & 24.62	& 0.7227	& 0.8667	& 0.1297	& 7.529\\
   \bottomrule[1pt]
\end{tabular}}
\vspace{-0.1cm}
\end{table}

\section{Loss functions}
\label{loss}
\vspace{-0.1cm}
The proposed framework obtains faithful light enhancement by optimizing the reconstruction loss, perceptual loss, and high-frequency loss.
We utilize three objective losses to optimize our network, including reconstruction loss ($L_{\textrm{Re}}$ and $L_{\text{\textrm{SSIM}}}$), perceptual loss ($L_{\mathrm{P}}$), and high-frequency loss ($L_{\mathrm{HF}}$). To summarize, the complete objective of our proposed model is combined as follows:
\vspace{-0.1cm}
\begin{equation}
L_{\mathrm{total}} = \alpha \cdot L_{\mathrm{Re}} + \beta \cdot L_{\mathrm{SSIM}} + \gamma \cdot L_{\mathrm{HF}}  + \eta \cdot L_{\mathrm{P}},
\vspace{-0.1cm}
\end{equation}
where $\alpha$, $\beta$, $\gamma$, and $\eta$ are the corresponding weight coefficients. 

\begin{figure}[htbp]
    \centering
    \includegraphics[width=0.98\textwidth]{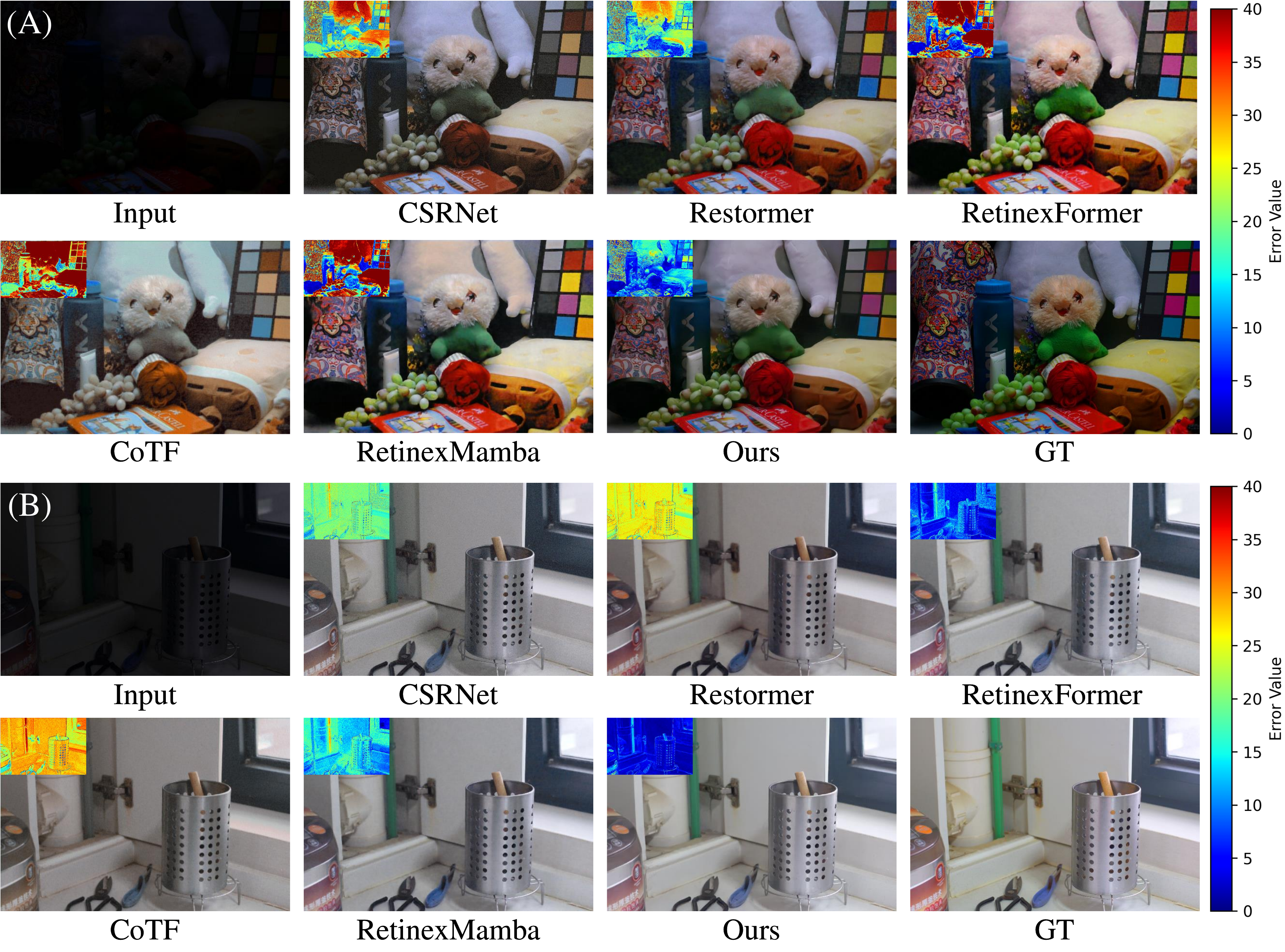}
    \vspace{-0.3cm}
    \caption{Visual comparisons between our LALet and the SOTA methods on the LOLv1 dataset.}
    \vspace{-0.3cm}
\label{lol}
\end{figure}

\end{document}